\documentclass[twocolumn,fleqn,10pt]{wlscirep}
\usepackage[utf8]{inputenc}
\usepackage[T1]{fontenc}
\usepackage{bm}
\usepackage{subcaption,booktabs}
\title{Multi-fidelity climate model parameterization for better generalization and extrapolation}

\author[1,*]{Mohamed Aziz Bhouri}
\author[2]{Liran Peng}
\author[2,3]{Michael S. Pritchard}
\author[1,4]{Pierre Gentine}
\affil[1]{Columbia University, Earth and Environmental Engineering, New York City, 10027, USA}
\affil[2]{University of California, Irvine, Center for Complex Biological Systems, Irvine, 92697, USA}
\affil[3]{NVIDIA, Santa Clara, CA, 95051, USA}
\affil[4]{Columbia University, Climate School, New York City, 10027, USA}

\affil[*]{mb4957@columbia.edu}

\begin{abstract} 

Machine-learning-based parameterizations (i.e. representation of sub-grid processes) of global climate models or turbulent simulations have recently been proposed as a powerful alternative to physical, but empirical, representations, offering a lower computational cost and higher accuracy. Yet, those approaches still suffer from a lack of generalization and extrapolation beyond the training data, which is however critical to projecting climate change or unobserved regimes of turbulence. Here we show that a multi-fidelity approach, which integrates datasets of different accuracy and abundance, can provide the best of both worlds: the capacity to extrapolate leveraging the physically-based parameterization and a higher accuracy using the machine-learning-based parameterizations. In an application to climate modeling, the multi-fidelity framework yields more accurate climate projections without requiring major increase in computational resources. Our multi-fidelity randomized prior networks (MF-RPNs) combine physical parameterization data as low-fidelity and storm-resolving historical run's data as high-fidelity. To extrapolate beyond the training data, the MF-RPNs are tested on high-fidelity warming scenarios, $+4K$, data. We show the MF-RPN's capacity to return much more skillful predictions compared to either low- or high-fidelity (historical data) simulations trained only on one regime while providing trustworthy uncertainty quantification across a wide range of scenarios. Our approach paves the way for the use of machine-learning based methods that can optimally leverage historical observations or high-fidelity simulations and extrapolate to unseen regimes such as climate change.

\end{abstract}
\begin{document}

\flushbottom
\maketitle
%
%
\thispagestyle{empty}


\section*{Introduction} 

Due to limited computational resources and the many scales required in climate  or turbulent simulations, unresolved sub-grid processes are approximated through parameterization schemes, or closures in numerical models. Parameterizations serve as approximate representations of small-scale processes and are the most dominant source of uncertainty in models predictions. To reduce these structural closures errors and uncertainties, several recent pieces of work have proposed machine-learning based parameterizations, which have been shown to dramatically improve the representation of physical processes and strongly reduce structural errors compared to standard schemes \cite{Brenowitz2018,Rasp2018,Gentine2018,OGorman2018,Bolton2019,Ross2023,Partee2022,Couvreux2021,Kashinath2021,Zanna2021,Sonnewald2021}. Another source of uncertainty stems from the inherent stochastic nature of many physical sub-grid processes in nature, such as turbulence or cloud micro-physics \cite{Lorenz96,Palmer2001,Tribbia2004,Karimi2010}. Stochastic parameterization schemes have been proposed to better characterize this latter source of uncertainty, as it can be important to correctly predict the prediction variability \cite{Palmer2012,Wang2016,Davini2017,Christensen2017,Strommen2018,Berner2012,Seiffert2010,Ajayamohan2013,Dawson2015,Berner2017}. 

Along with the development of recent climate parameterization schemes, various climate simulation data have been made available. However, most of these were built with simple aqua-planets \cite{Gentine2018, Rasp2018, Brenowitz2020, Han2020, Ott2020, Iglesias2023} and those that considered real geography \cite{Mooers2021, Wang2022a} did not include enough variables for a complete land-surface coupling. Hence, there is a wealth of relatively low-fidelity climate simulation data that is available to build climate parameterization schemes, while high-fidelity datasets based on high-resolution and/or multi-scale climate simulations are rare. Therefore, there is a clear need to investigate the possibility of building schemes that take advantage of the abundant low-fidelity data in order to improve high-fidelity parameterizations. In addition, although several machine learning-based methods have been successfully developed in order to parameterize turbulence \cite{Frezat2022}, atmospheric \cite{Krasnopolsky2005,Schneider2017,Brenowitz2018,Rasp2018,Gentine2018,OGorman2018,Mooers2021,Gettelman2021} and oceanic processes \cite{Bolton2019}, these methods struggle with out-of sample testing inputs and are unable to extrapolate beyond the training data regimes and scenarios \cite{Rasp2018} (out-of-distribution limitations). An important body of recent work has made exciting progress on using machine learning methods to reduce biases in climate simulations \cite{Bretherton2022, Clark2022, Kwa2023, Sanford2023}. However, these approaches were restricted to improving coarse-grid climate and weather models using higher resolution simulations while the opposite would of greater use given the abundant low-fidelity data.

Multi-fidelity (MF) models have recently been successful in several computational science and engineering applications \cite{Godino2023,Perdikaris2017,Meng2021,Chen2021,Liu2019,Zhang2021,Wu2022}. These models are suitable for problems where multiple datasets or computational models are available for a given system of interest. MF models aggregate data and information with different fidelity, i.e. level of accuracy and details availability \cite{Peherstorfer2018}. High-fidelity (HF) models or datasets provide more accurate information but require greater computational or measurement resources. On the other hand, low-fidelity (LF) models or datasets are less accurate but cheaper to run or obtain, and hence generally more abundant compared to HF simulation runs or data \cite{Godino2017}. 

In this work we use a probabilistic MF approach in order to allow uncertainty quantification. Different Bayesian models can be used in order to build MF approaches including: Markov-Chain Monte Carlo (MCMC) sampling methods \cite{Neal2011}, variational inference techniques \cite{Hinton1993,Blundell2015}, deep ensembles \cite{Lakshmin2017,Fort2019} and dropout \cite{Srivastava2014,Gal2016}. Given the typical dimensionality and size of the datasets for Earth System Model (ESM) parameterizations, the gold-standard MCMC methods are out of scope. Besides, variational inference approximations can suffer from posterior variance underestimation and results in a poor approximation of the true multi-modal posterior distribution when applied to deep learning frameworks \cite{Osband2022}. In addition, it has been shown that dropout and standard deep ensemble methods often provide minimal uncertainty estimates which prevents their use in applications requiring sufficiently accurate approximation of posterior distributions \cite{Riquelme2018,Osband2022}. 

Randomized Prior Networks (RPNs) \cite{Osband2018} were developed in order to provide a compromise between acceptable computational cost for building Bayesian surrogate models and overcome the uncertainty underestimation. RPNs take advantage of an explicit incorporation of prior knowledge in order to improve the model predictions in regions where limited or no training data is available \cite{Osband2018,Yang2020,Bhouri2023}. There has been additional theoretical studies proving the conservative uncertainty obtained with RPNs and their ability to reliably detect out-of-distribution samples \cite{Ciosek2020}. RPNs have also been proven to outperform HMC methods, variational inference techniques and dropout as a Bayesian approximation in the context of complex sequential decision making tasks \cite{Ciosek2020,Osband2022}. The RPNs improvement is mainly driven by their parallelizable implementation resulting in a significantly lower computational cost and the possibility of building Bayesian surrogate models for complex and large neural network architectures. 

Extending on previous deterministic neural network parameterization studies of atmospheric ESM parameterization \cite{Mooers2021}, here we propose a multi-fidelity RPN model (MF-RPN) as a parameterization scheme for atmospheric convection (deep clouds), which is the first of its kind to the best of our knowledge. The MF-RPN surrogate model is designed to take into account the distribution shift across regimes leveraging the rich LF training data regimes while refining it with higher accuracy but more limited HF regimes. This proves  crucial to obtain skillful extrapolation predictions for unseen HF testing data. We show that the proposed MF-RPN can provide the best of both worlds: the higher accuracy of the HF data and the generalization capability thanks to the LF one. The improved MF-RPN skillful predictions are tested across various error metrics on HF data of unseen warmer climate scenarios and against three other surrogate models.

\begin{figure*}
     \centering
     \includegraphics[width=0.6\textwidth]{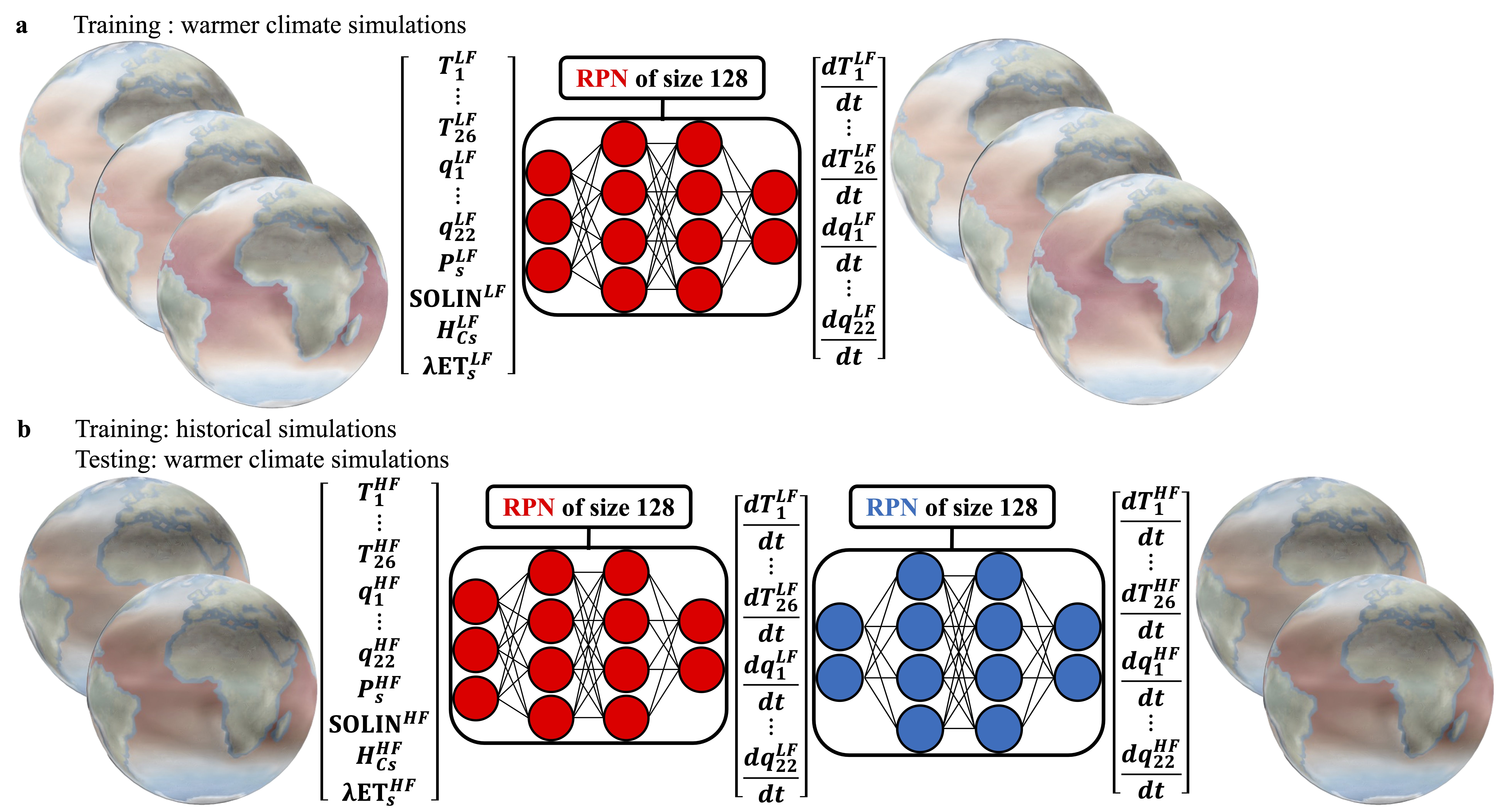}
         \caption{\textbf{Multi-fidelity problem setting for ESM parameterization}. $T_i$, $i=1,\ldots,26$ refer to atmospheric temperature $[\mathrm{K}]$ at different vertical levels. $q_i$, $i=1,\ldots,22$ refer to specific humidity $[\mathrm{kg/kg}]$ at different vertical levels.  $P_s$, $\mathrm{SOLIN}$, $H_{Cs}$ and $\mathrm{\lambda ET}_s$ refer to surface pressure $[\mathrm{Pa}]$, TOA solar insulation $[\mathrm{W/m^2}]$, sensible heat flux $[\mathrm{W/m^2}]$ and latent heat flux $[\mathrm{W/m^2}]$ respectively. Parameterization input and output are of dimension $52$ and $48$ respectively. \textbf{a}, Low-fidelity problem setting for multi-fidelity RPN-based CAM5/SPCAM5 convection superparameterization. \textbf{b}, High-fidelity problem setting for multi-fidelity RPN-based CAM5/SPCAM5 convection superparameterization}
     \label{fig:overview}
\end{figure*}

\section*{Results} 

\subsection*{Problem setup}
We define the convection superparameterization problem considered and the used ESM datasets.

\subsubsection*{ESM convection superparameterization}
The problem considered here consists of predicting subgrid-scale tendencies of heat and moisture convection (i.e. time rate of change) at all vertical levels and for every timestep \cite{Mooers2021}. The convection parameterization for climate models is becoming a mature problem given the recent studies focusing on it \cite{Gentine2018,Rasp2018,Brenowitz2020,Behrens2022,Yu2023}. The parameterization input is similar to the standard Community Earth System Model version 2.1.3 (CESM2.1.3) Community Atmospheric Model version 5 (CAM5) parameterization and is taken as coarse-grid atmospheric thermodynamics components consisting of: atmospheric temperature for each of the $26$ vertical levels spanning the column and specific humidity for each of the $22$ vertical levels spanning the column except the first four levels from top of atmosphere (TOA) (Methods). The input vector also contains the surface pressure, TOA solar insulation, surface latent heat flux and surface sensible heat flux (figure \ref{fig:overview}).

The parameterization output is the subgrid-scale convective tendency of temperature, or heat tendency for short, and the subgrid-scale convective tendency of specific humidity throughout the column, or moisture tendency for short (figure \ref{fig:overview}). This tendency definition accounts for the sub-grid advection of temperature and moisture by convection and fine-scale turbulence, as well as for the effect of radiative heating throughout the column on temperature tendency.

\subsubsection*{Datasets}


Our evaluation consists of the CESM (high-fidelity) Super-parameterized Community Atmospheric Model version 5 (SPCAM5) in a real geography setup (Methods). Unlike CAM5, SPCAM5 nearly explicitly  resolves atmospheric moist convection (including deep convection) by using idealized embedded cloud resolving models \cite{Grabowski2001,Randall2003}, introducing less physical approximations. Both models incorporate the CAM radiation package (CAM-RT) \cite{Iacono2008,Mlawer1997}. The HF training data, of size $29.5$ M points (pixels x times), is constructed by considering the SPCAM5 historical run simulation (i.e. non-global warming) of three months (Methods). 

The proposed multi-fidelity model aggregates high-fidelity SPCAM5 historical dataset and low-fidelity (historical and future) CAM5 data. SPCAM5 has a much higher computational cost and better capability to resolve convection compared to CAM5. CAM5 uses a physically-based parameterization of convection, introducing more physical approximations compared to SPCAM5, which resolves deep convection. In addition, radiation in CAM5 is estimated every 2 time-steps, while the radiation is estimated every 1 time-step for SPCAM5 by default. Therefore, CAM5 is a good candidate for a low-fidelity model of SPCAM5, as it is computationally cheaper but also less accurate. For all CAM5 and SPCAM5 simulations, the cosine of the solar zenith angle is estimated as a function of Julian calendar day, latitude, longitude and Solar declination.

Since we are interested in warmer climate scenarios, the CAM5 
simulation corresponding to a global warming situation, with a prescribed sea surface temperature (SST) that has been augmented by $4$ K and $8$ K, referred to as $+4$K and $+8$K simulations respectively, were considered as LF data candidates for training. A comparison of their inputs and outputs' distributions with those of the SPCAM5 training data shows a more pronounced extrapolation regime for CAM5 $+8$K data, mainly due to the increased holding capacity of moisture in the atmosphere with climate change (Clausius-Clapeyron) at higher temperatures. Hence, the CAM5 $+8$K was chosen as low-fidelity model for extrapolation (Methods).

Since we are interested in extrapolating beyond the training data, the test data is constructed by considering the CESM SPCAM5 $+4$K simulation. In order to enhance the extrapolation evaluation to unseen data, the testing dataset corresponds to a full year of the $+4$K simulation, resulting in a final test dataset of roughly $121.1$ M points (Methods). Beyond global warming, the test dataset also extrapolates to unseen phases of the SPCAM5 seasonal cycle.

\subsection*{Surrogat models}

We provide description of the four surrogate models that are built for the CAM5/SPCAM5 convection superparameterization.

\subsubsection*{Single-fidelity Randomized Prior Networks}

In this work, Bayesian models are constructed using an ensemble method called Randomized Prior Networks (RPNs) \cite{Osband2018}. Each member of the RPNs is built as the sum of a trainable and a non-trainable (so-called ``prior'') surrogate model; we used fully-connected neural network for simplicity. Multiple replicas of the networks are constructed by independent and random sampling of both trainable and non-trainable parameters \cite{Yang2022, Bhouri2023}. The non-trainable parameters are initialized but then kept fixed during the fitting process which only optimizes over the trainable parameters. In our case of fully-connected neural networks, we resort to Glorot initialization \cite{Glorot2010}, which defines the probability distributions from which the fixed non-trainable parameters are sampled. RPNs also resort to data bootstrapping in order to mitigate a potential uncertainty collapse of the ensemble method when tested beyond the training data points \cite{Bhouri2023}. Data bootstrapping consists of sub-sampling and randomization of the data on which each network in the ensemble is trained.

The Single High Fidelity model corresponds to a standard RPN trained only on the HF data and will be referred to as SF-HF-RPN.  Hyperparameters of individual neural networks did not need to be tuned from scratch. They were instead chosen based on the hyperparameter optimization over $\sim 250$ trials conducted in \textit{Mooers et al.}'s study on fully-connected neural network convection superparameterization for SPCAM5 \cite{Mooers2021} (Methods). RPN ensembles of 128 networks were considered as justified in \textit{Yang et al.} \cite{Yang2022}.

\subsubsection*{Deterministic neural network}
In addition to the SF-HF model, we also considered, for reference, a deterministic model defined as a single fully-connected neural network with the same hyperparameters as the SF-HF model's individual neural networks. Both deterministic and SF-HF-RPN models were trained on the SPCAM5 HF historical run data providing a baseline for high-fidelity models trained only on historical data.

\subsubsection*{Multi-fidelity Randomized Prior Networks}

The multi-fidelity model is also constructed using RPNs of size 128. Trainable and non-trainable surrogate models of each member of the multi-fidelity RPN (MF-RPN) are built with the architecture detailed in figure \ref{fig:overview}.b. The chosen architecture consists of two fully connected deep neural networks. The first network (highlighted in red in figure \ref{fig:overview}) predicts the low-fidelity parameterization output from the parameterization input, while the second network (highlighted in blue in figure \ref{fig:overview}) predicts the high-fidelity parameterization output as a function of the low-fidelity parameterization output. The trainable surrogate model of each member of the MF-RPN is trained using a joint training of both networks (Methods). 

Our MF-RPN learns the mappings between related physical variables: emulating the parameterization (inputs to outputs) at low-fidelity (red network in figure \ref{fig:overview}), and mapping the parameterization outputs at different fidelity levels (low to high-fidelity, blue network in figure \ref{fig:overview}.b). The proposed architecture directly learns the non-linear mapping between the low-to-high fidelity outputs instead of inferring the difference between them as an error bias correction \cite{Kwa2022,Clark2022,WattMeyer2022,Kwa2023}. The bias correction approach was only shown to improve coarse-grid climate models using higher resolution simulations and not vice versa despite the abundant low-fidelity data. In addition, the chosen architecture naturally accommodates outputs of different dimensions for different fidelity levels. In the case of inputs of different dimensions for different fidelity levels, an additional neural network can be added in order to infer the mapping between the different inputs. Besides, the chosen architecture naturally ensures uncertainty propagation between different fidelity levels since low-fidelity predictions are directly fed as inputs for the high-fidelity model within the MF-RPN (Methods). Finally, since the low-fidelity training data was built such that it provides the MF model with useful information regarding the high-fidelity extrapolation scenarios, the MF-RPN model is trained on normalized data with respect to the statistics of the CAM5 $+8K$ run data in order to take into account the data distribution shift between different fidelity levels (Methods).


\subsubsection*{Low-fidelity Randomized Prior Networks}

A low-fidelity RPN model can be considered based on the MF-RPN model detailed above without any further training. Indeed, the low-fidelity network within the MF-RPN model (red network in figure \ref{fig:overview}) already provides predictions for the convection parameterization outputs. Hence we can also test this LF-RPN model on high-fidelity data points by considering the corresponding parameterization inputs. The LF-RPN can be seen as a control model whose performance allows assessing whether the MF-RPN model is capable of properly aggregating both datasets to well generalize beyond the training data. If both models performance are similar, then the MF-RPN improvement would solely be due to being trained on the abundant low-fidelity data for a warmer climate and with a full seasonal cycle. However, if the MF-RPN results improves upon the LF-RPN ones, then it would justify that the MF-RPN model is well capable of merging both training datasets, including the scarce but more physically sound high-fidelity data even without the full seasonal cycle. 

\subsection*{Forecast skills}

All surrogate models are evaluated based on their performance on the high-fidelity test dataset corresponding to the SPCAM5 $+4K$ simulation. 

\subsubsection*{Evaluation metrics}

Different evaluation metrics are considered and computed for each output variable. We report the mean absolute error (MAE) and the coefficient of determination ($R^2$). The MAE is always positive and a lower value corresponds to a more accurate model. The coefficient of determination is upper-bounded by $1$ and values closer to $1$ correspond to more accurate models (Methods).

\subsubsection*{Forecast skills results}

For the heat tendency, the MF-RPN is the only model with  positive global $R^2$ values for all vertical levels, with an average $R^2$ of $0.62$ across all levels (figure \ref{fig:glob_errs}.a and figure \ref{fig:glob_R_CRPS_grey_with_negative}.a in Supplementary Information showing the negative values for $R^2$ where appropriate). Besides, MF-RPN is always the best model except for the the $137$ and $160 \mathrm{hPa}$ vertical levels. Except for the lowest and highest vertical levels, LF-RPN is outperforming both deterministic NN and SF-HF-RPN models (figure \ref{fig:glob_errs}.a). Hence, for these two levels, historical SPCAM5 simulations are closer to those of SPCAM5 $+4K$ run. However, for any other vertical level except the first one at TOA and the closest one to sea surface, CAM5 $+8K$ simulation provides a better approximate of SPCAM5 $+4K$ run dynamics. For most of the vertical levels beyond the two extreme ones, MF-RPN is improving upon LF-RPN which in turn is outperforming deterministic NN and SF-HF-RPN models. In addition, for lowest and highest vertical levels, MF-RPN is improving upon the deterministic NN and SF-HF-RPN models which in turn are outperforming the LF-RPN model. Hence, the MF-RPN has the ability to get the best of the both worlds by aggregating both datasets of different fidelity levels as (1) it learns from a high-fidelity parameterization that resolves convection based on the high-fidelity dataset and (2) generalizes beyond the high-fidelity training data regime thanks to the informative low-fidelity simulations covering regimes at higher sea surface temperatures. It is worth noting that the SF-HF-RPN is capable of improving upon the deterministic NN for nearly all vertical levels and even for determinstic error metrics (figure \ref{fig:glob_errs}.a), showing the benefits of using RPNs as a stochasticity-aware surrogate model.

For the moisture tendency, the overall performance of all models for almost all vertical levels is lower in terms of $R^2$ compared to the results obtained for the heat tendency (figure \ref{fig:glob_errs}). This result can be mainly attributed to the higher stochasticity of humidity and precipitation compared to the temperature. The MF-RPN model is the best performing model for all pressure levels except for the $188 \mathrm{hPa}$ vertical level where the LF-RPN is the best one, and for the closest level to the surface ($958 \mathrm{hPa}$) where MF-RPN is outperformed by the deterministic NN and SF-HF-RPN (figure \ref{fig:glob_errs}.b). It is worth noting that for this level, MF-RPN is still performing well with an $R^2=0.71$, unlike LF-RPN showing a negative $R^2=-0.6$ (figure \ref{fig:glob_R_CRPS_grey_with_negative}.b). For all pressure levels where moisture tendency is the most significant and critical for cloud formation (typically between $250$ and $750 \mathrm{hPa}$), the MF-RPN model clearly outperforms all other models with an average $R^2$ equal to $0.73$ across different vertical levels (figure \ref{fig:glob_errs}.b). In addition, for all levels where the deterministic NN, SF-HF-RPN and LF-RPN all fail (e.g. all levels below $160 \mathrm{hPa}$, $897$ and $937 \mathrm{hPa}$), the MF-RPN is still capable of providing significantly better results than all these models showing even positive $R^2$ values (e.g. $0.36$ and $0.32$ for levels $897$ and $937 \mathrm{hPa}$) thanks to both datasets aggregation.

The LF-RPN is outperforming the determinisitic NN. and SF-HF-RPN models except within the stratosphere (where convection is absent anyways) and for pressure levels close to the surface (figure \ref{fig:glob_errs}.b). This result confirms that within the highest and lowest vertical levels, historical SPCAM5 simulation dynamics are closer to those of SPCAM5 $+4K$ run, while beyond them CAM5 $+8K$ simulation provides a better approximate of SPCAM5 $+4K$ run dynamics. Finally, for most of vertical levels from TOA to $494 \mathrm{hPa}$, the deterministic NN. is outperforming the SF-HF-RPN, while the opposite is observed for all vertical levels from $581$ to $958 \mathrm{hPa}$. Hence, the SF-HF-RPN is only capable of better resolving the moisture convection stochasticity for vertical levels below the $494 \mathrm{hPa}$ one, while it struggles to do so at higher levels.


\begin{figure*}[ht]
\centering
\includegraphics[width=0.6\linewidth]{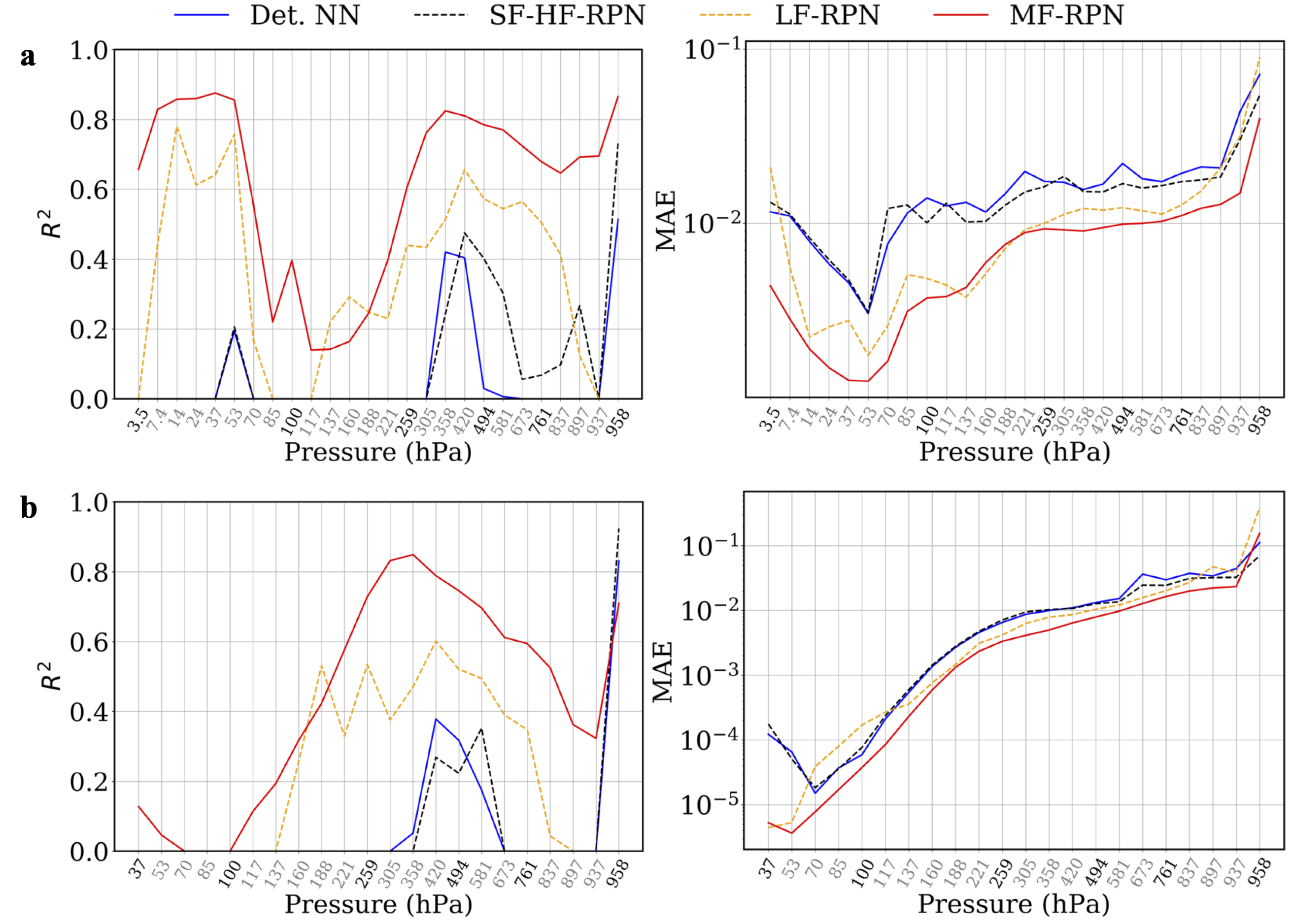}
\caption{\textbf{$\mathbf{R^2}$ and MAE metrics evaluation for different models across all test data points concatenated over space and time}. Negative $R^2$ values are lumped to $0$ for clarity purposes. \textbf{a}, Heat tendency results. \textbf{b}, Moisture tendency results}
\label{fig:glob_errs}
\end{figure*}

The SF-HF-RPN model has higher $R^2$ values for the moisture tendency than the deterministic NN in the temperate zone thanks to its capacity of better resolving the moisture convection stochasticity within this region (figure \ref{fig:moist_rgriff_0_1_spatial_18} and figure \label{fig:long_lat_moist_all} in Supplementary Information). However, the SF-HF-RPN model fails to provide more accurate predictions within the tropics and polar regions. The LF-RPN model improves further upon the SF-HF-RPN model within the temperate zone and even within the tropics and polar regions. These results confirm the informative capacity of the low-fidelity data for the extrapolation scenario of interest and also the LF-RPN capacity to resolve the moisture convection stochasticity since it is an ensemble method. Finally, the MF-RPN model improves even further upon the LF-RPN model across all regions with a nearly perfect $R^2$ score in the temperate zone. The MF-RPN model also shows better results for all tropical regions (figure \ref{fig:moist_rgriff_0_1_spatial_18} and figure \label{fig:long_lat_moist_all} in Supplementary Information). This result is of a significant importance since we are extrapolating to warmer climates and hence the tropics (the warmest region of the world) provide test data-points that are well outside the training datasets distributions. In addition, the tropics is a challenging region to model in terms of convection and ESMs exhibit many typical problems within this region that are related to sub-grid convection parameterizations. Among these problems we can mention the double inter-tropical convergence zone (ITCZ) \cite{Oueslati2015}, too much drizzle and missing precipitation extremes \cite{Bador2020}, and an unrealistic equatorial wave spectrum with a missing Madden–Julian oscillation (MJO) \cite{Meng2013}. Therefore, providing a framework to improve convection paramterization within this region can help remedy these issues. 

\begin{figure*}
\centering
\includegraphics[width=0.7\linewidth]{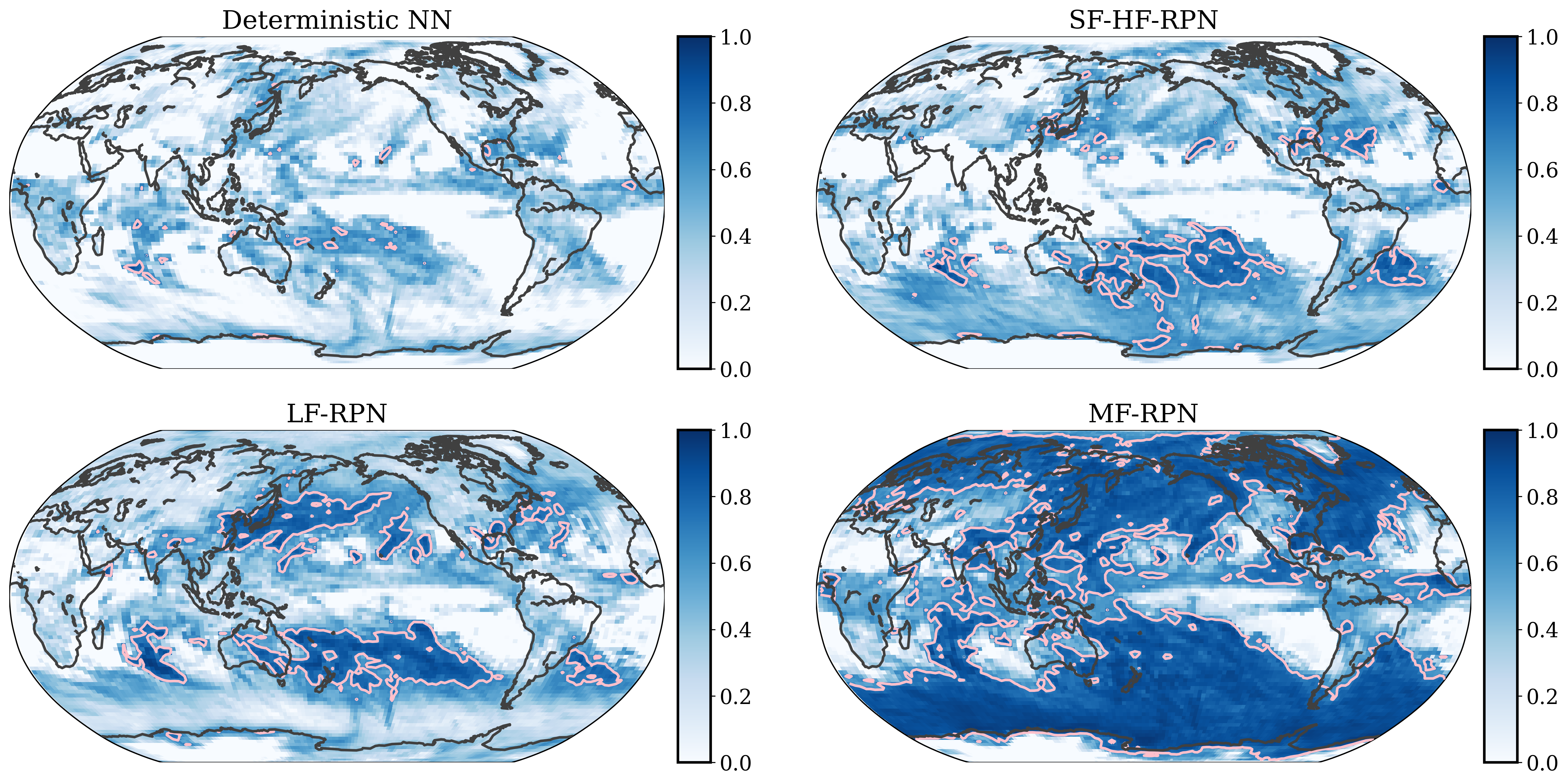}
\caption{\textbf{Longitude-latitude variation of coefficient of determination $\mathbf{R^2}$ for moisture tendency at vertical level $\mathbf{ P = 494 \ \mathrm{\mathbf{hPa}} }$ for different surrogate models}. $\mathbf{R^2}$ is evaluated on the test dataset and negative values are lumped to $0$ for clarity purposes.}
\label{fig:moist_rgriff_0_1_spatial_18}
\end{figure*}

In Supplementary Information, we provide all longitude-latitude variations of the MAE and $R^2$ metrics for heat and moisture tendencies for pressure levels: $259$, $494$ and $761 \ \mathrm{hPa}$ (figures \ref{fig:long_lat_moist_all} and \ref{fig:long_lat_heat_all}). These results show a very similar behavior as observed above for the moisture tendency at level $494 \ \mathrm{hPa}$. For other levels and for heat tendency, the MF-RPN model shows even higher $R^2$ values in the south Atlantic ocean and African Sahara. We also provide the temporal variation of the MAE and $R^2$ metrics in Supplementary Information.

\begin{figure*}
\centering
\includegraphics[width=0.7\linewidth]{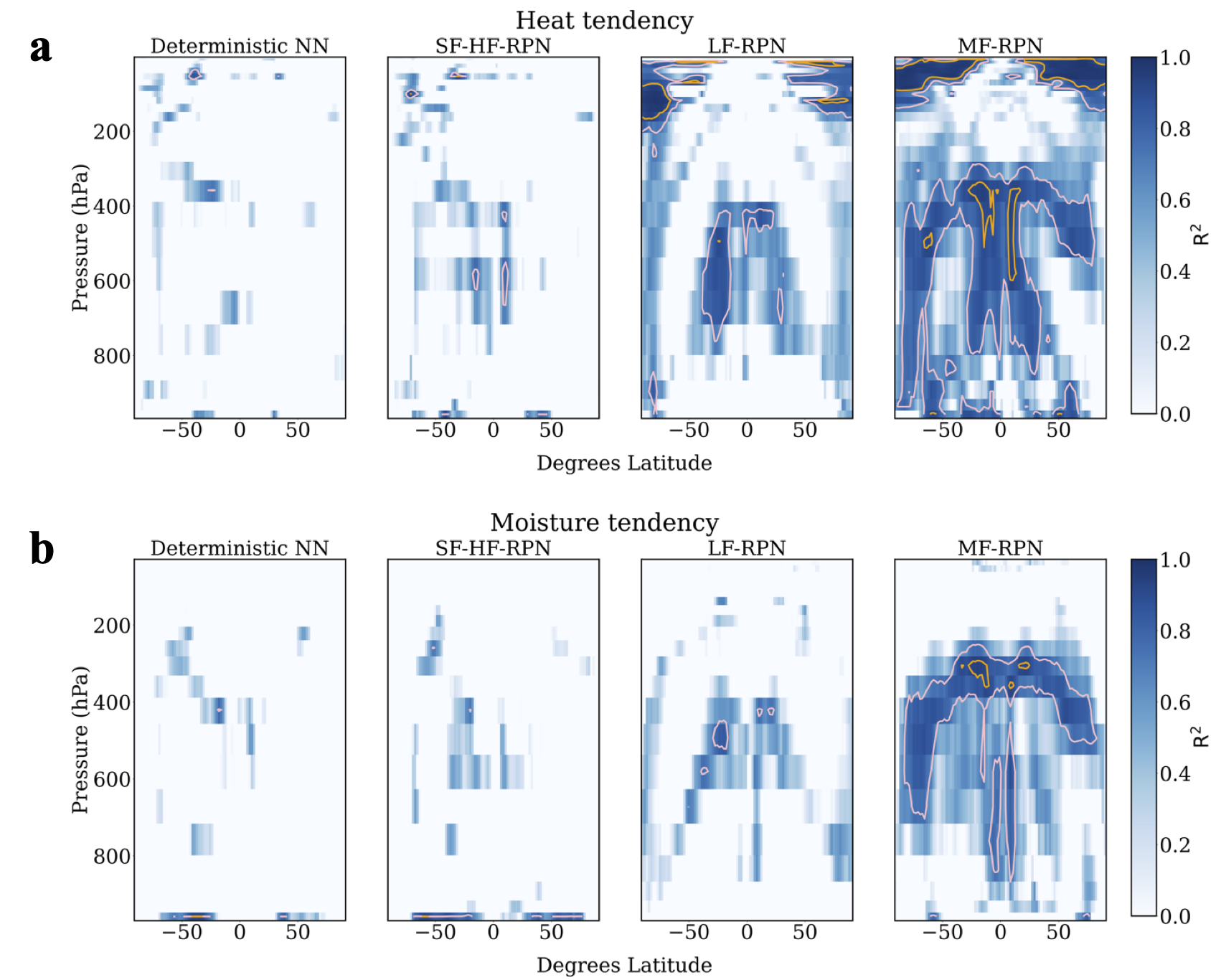}
\caption{\textbf{Pressure-latitude variation of coefficient of determination $\mathbf{R^2}$ for different surrogate models}. $\mathbf{R^2}$ is evaluated on the test dataset and negative values are lumped to $0$ for clarity purposes. \textbf{a}, Heat tendency results. \textbf{b}, Moisture tendency results.}
\label{fig:press_lat_heat_moist_rgriff}
\end{figure*}

For both tendencies and nearly all vertical levels, the MF-RPN model shows improved results compared to all other surrogate models including for the tropics across all vertical levels between around $250$ and $800 \ \mathrm{hPa}$, where lies the double ITCZ region (figures \ref{fig:press_lat_heat_moist_rgriff}.a and \ref{fig:press_lat_heat_moist_rgriff}.b). The MF-RPN shows significant improvement for the heat tendency parameterization in the stratosphere, mostly within the polar region and the temperate zone. It also displays a better parameterization for the heat tendency within the first vertical level close to sea surface, showing a better parameterization for boundary layer regions (figure \ref{fig:press_lat_heat_moist_rgriff}.a). For the heat tendency, the SF-HF-RPN model improves upon the deterministic NN. mostly within the tropics thanks to a better stochasticity representation (figure \ref{fig:press_lat_heat_moist_rgriff}.a). However, the improvement is only noticeable between the $250 \ \mathrm{hPa}$ and $750 \ \mathrm{hPa}$ pressure levels, which are the critical levels for cloud formation. The LF-RPN model improves further compared to the SF-HF-RPN model for the tropics between the $250 \ \mathrm{hPa}$ and $750 \ \mathrm{hPa}$ pressure levels, and also shows higher $R^2$ values for both polar regions, including a very pronounced improvement in these regions within the stratosphere (figure \ref{fig:press_lat_heat_moist_rgriff}.a). Compared to the LF-RPN, the MF-RPN model improves the heat tendency parameterization results for the south pole across nearly all pressure levels, while it under-performs in the north pole for vertical levels below $300 \ \mathrm{hPa}$. 

For the moisture tendency, the SF-HF-RPN still shows some improvement compared to the deterministic NN model, mostly within the southern temperate zone (figure \ref{fig:press_lat_heat_moist_rgriff}.b). The LF-RPN model improves further compared to the SF-HF-RPN model for the tropics between the $400$ and $600 \ \mathrm{hPa}$ pressure levels, which is a smaller region compared to the improvement observed for the heat tendency. The LF-RPN model also shows higher $R^2$ values for both polar regions. The MF-RPN model improves further upon the LF-RPN in the tropical region between the $200$ and $900 \ \mathrm{hPa}$ vertical levels, and also in the temperate zone between $250$ and $700 \ \mathrm{hPa}$ levels for the south hemisphere (figure \ref{fig:press_lat_heat_moist_rgriff}.b). The temperate zone improvement applies to a smaller region mostly located between $250$ and $550 \ \mathrm{hPa}$ levels for the north hemisphere. This observation is coherent with the heat tendency results showing a higher MF-RPN's performance for the temperate zone in the southern hemisphere compared to the northern one.

Finally, we verify that the MF-RPN's uncertainty quantification estimated over the ensemble predictions is coherent as it increases with the predictions error (figure \ref{fig:uncert_density_all_models} in Supplementary Information). This means that without accessing any information on the true target values, the MF-RPN model is intrinsically capable of estimating its predictions accuracy across different test data points. We also verify that the longitude-latitude structure of the uncertainty well matches with the longitude-latitude variation of the predictions error, with the highest values being observed around the tropics where the inherent stochasticity of convection is the highest compared to other regions (figure \ref{fig:uncert_long_lat_all}). 

\section*{Discussion}


Extrapolation beyond training datasets is a long-standing problem of importance for machine-learning-based models, and for the emulation of physical models in particular. In this work, we showed how the proposed multifidelity (here with an RPN) approach can tackle this problem by considering the high-fidelity convection data on historical observations and optimally merging it with a prior coming from a physically based parameterization exploring more diverse regimes as it is computationally cheaper. We showed that the proposed approach can extrapolate heat and moisture convection predictions over substantial climate warming situations, where existing supervised (single-fidelity) methods struggle. The improvement includes even the tropics where convection stochasticity is higher compared to other regions and where different Earth system models exhibit many typical problems related to sub-grid convection parameterizations. We also verified that the proposed multifidelity-RPN uncertainty quantification coherently increases with predictions error. The proposed MF parameterization approach can also be combined with explainable AI techniques to further study similarities and discrepancies between different Earth system models.

The multifidelity-RPN performance is due to the model's design (architecture accounting for data distribution shift) and to its optimal aggregation of different datasets of different fidelity levels. This latter property allows the MF model to provide the best of both worlds: the capacity to extrapolate based on low-fidelity data exploring many regimes of convection and the higher accuracy based on high-fidelity data covering more limited regimes because of its computational cost. Hence, the MF-based parameterization narrows further the gap between the climate science and machine-learning communities by (1) building trust in the capacity of ML-based parameterization to extrapolate to unknown scenarios thanks to its low-fidelity component, (2) while also harnessing more physically-consistent and higher accuracy high-fidelity data.

There is still room for improving the proposed multifidelity parameterization scheme by enforcing physical constraints \cite{Beucler2021a, Reed2023}, aggregating observational data and extending it to an online setting within differentiable solvers when available \cite{Frezat2022,Bhouri2023b}. Nonetheless, whereas existing machine learning-based climate parameterizations struggle to generalize beyond the training data regimes, we hope that thanks to the multifidelity extrapolation capabilities, this work will pave the way to finally tackle climate change projection with Artificial Intelligence.

\newpage 

\section*{Methods} 
\label{sec:methods}
\subsection*{Earth system model convection superparameterization}

Our base model is the CESM2.1.3 CAM5 model with real-geography boundary conditions. CAM5 uses a physically-based parameterization of convection and is hence taken as low-fidelity model. The high-fidelity model is taken as the super-parameterized CAM version 5 model (SPCAM5). Notably, while CAM5 employs certain standard packages, SPCAM5 distinguishes itself by its capability to explicitly resolve sub-grid scale physical processes, making it computationally broader in scope \cite{Yi2016}. Within CAM5, the micro-physics is driven by the two-moment bulk strati-form cloud micro-physics scheme \cite{Morrison2008}. CAM5 macro-physics draws from Park and Bretherton's shallow convection and moist turbulence schemes \cite{Park2009}, and its planetary boundary layer (PBL) packages are based on Bretherton and Park' moist turbulence parameterization \cite{Bretherton2009}. SPCAM5 uses idealized cloud resolving models (CRM) in order to nearly explicitly  resolve atmospheric moist convection. In particular, SPCAM5 uses the one-moment cloud micro-physics. The SPCAM5 runs considered use 32 CRM columns and 25 CRM vertical levels. 

For SPCAM5 training and testing datasets considered, the first $4$ vertical levels starting form Top Of the Atmosphere (TOA) show all zero values for the moisture tendency across all earth and for the whole simulations time periods. Therefore, the first $4$ vertical levels starting form TOA have been discarded for the moisture tendency in the parameterization problem, and coherently for the specific humidity. 

\subsection*{Datasets}
\label{ss:LF_data}

All CAM5 and SPCAM5 simulations considered commence using climatological input data derived from a 20-year mean span around the year 2000. This data includes relevant solar radiation, greenhouse gas levels, oxidant concentrations, and present-day aerosol emissions (denoted as F2000). The prescribed SST and sea ice data sets were constructed as a blended product, using the global HadISST OI data set \cite{Hurrell2008}. The considered forcing consists of annually repeating climatological SSTs with full seasonality. In the simulations labeled $+4$K and $+8$K, the standard SST is elevated by $4$K and $8$K respectively.

The high-fidelity SPCAM5 training data is constructed by considering a historical run simulation while allowing for a model spin-up of a month. The training data corresponds to the time period from February 1st 2003 to April 31st 2003. The horizontal grid resolution of the ESM consists of a $1.9^\circ \times 2.5^\circ$ finite-volume dynamical core (i.e., $13824$ grid cells with $96$ in latitude and $144$ in longitude). The vertical resolution varies from $\approx 150 \ \text{m}$ to $\approx 5300 \ \text{m}$. The ESM time step is $30$ min and a temporal sub-sampling by a factor of 2 is performed (to reduce the overly correlated training data), resulting in a final training dataset of roughly $29.5$ M points.

For the high-fidelity SPCAM5 testing data, the corresponding temporal and spatial resolutions, model spin-up and temporal sub-sampling are the same as detailed above for the historical run. The testing dataset corresponds to a full year of the $+4$K simulation, covering  the time period from February 1st 2003 to January 31st 2004, resulting in a final test dataset of roughly $121.1$ M points. The testing dataset is constructed with a full-year simulation in order to have a comprehensive analysis of the models performance when tested on unseen climate scenarios and extrapolated to other phases of the SPCAM5 seasonal cycle. 

\begin{figure}[ht]
\centering
\includegraphics[width=\linewidth]{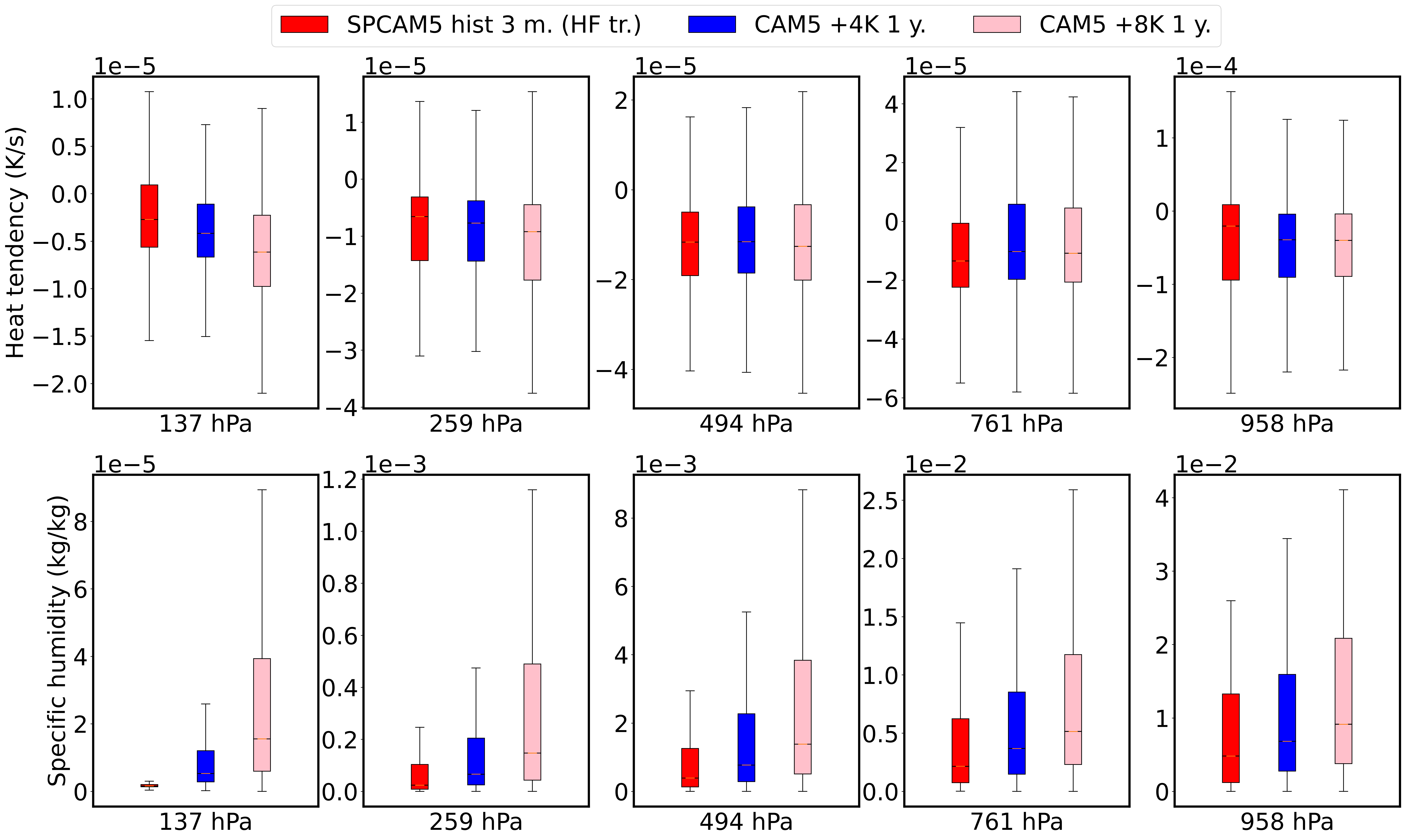}
\caption{Data distribution of the specific humidity and heat tendency for SPCAM5 training data (historical simulation) and two potential CAM5 training datasets ($+4$K and $+8$K simulations) at 5 different vertical levels.}
\label{fig:candle_plots_final}
\end{figure}

Given the testing dataset defined above, a straightforward choice for the low-fidelity CAM5 training dataset would be to consider a $+4$K simulation as defined for SPCAM5 for testing. However, low-fidelity data should not be defined with the assumption of prior knowledge of the testing data, but rather on the exploration of scenarios and regimes it provides beyond those observed within the high-fidelity training data. Hence, both CAM5 $+4$K and $+8$K are considered as potential candidates. The corresponding temporal and spatial resolutions, model spin-up and temporal sub-sampling are the same as detailed above for SPCAM5 simulations. In the context of multi-fidelity modelling and given the lower CAM5 computational cost, the simulation time period was taken from  February 1st 2003 to January 31st 2004 in each simulation. An analysis of the inputs and outputs' distributions for CAM5 $+4$K and $+8$K training datasets shows a broader distribution for the CAM5 $+8$K specific humidity across all pressure levels considered compared to the CAM5 $+4$K dataset (figure \ref{fig:candle_plots_final}). Hence CAM5 $+8$K provides a broader extrapolation regime due to the increased holding capacity of moisture in the atmosphere with climate change (Clausius-Clapeyron). CAM5 $+8$K also provides a clearer extrapolation for the heating tendency than CAM5 $+4$K when compared to the high-fidelity SPCAM5 historical run simulation (figure \ref{fig:candle_plots_final}). Based on the data distribution comparison, the CAM5 $+8$K is chosen as low-fidelity model since it provides a significantly more pronounced extrapolation beyond the regimes spanned by the SPCAM5 training dataset compared to CAM5 $+4$K. This property proves being crucial in obtaining skillful extrapolation predictions when the multi-fidelity model is tested on unseen SPCAM5 $+4$K data.  


\subsection*{Multi-fidelity Randomized Prior Networks}


The trainable surrogate model of each member $j$ of the MF-RPN is fitted using a joint training strategy of both networks. Hence the corresponding loss function that is minimized by stochastic gradient descent contains two terms. One term ensures that the first network learns the low-fidelity parameterization (red network in figure \ref{fig:overview}.a). The second term ensures that the pipeline through both networks learns the high-fidelity parameterization (red and blue networks in figure \ref{fig:overview}.b). Let $f_{\theta_{LF},j}$ denote the red network learning the LF paramterization, and $f_{\theta_{HF},j}$ the blue one learning the mapping to the HF parameterization output. These two networks are trained jointly via the minimization of the following loss function:
\begin{multline}
\mathcal{L} = \frac{1}{N_L} \sum_{i=1}^{N_L} \Big( y_{LF,i}-f_{\theta_{LF},j}(x_{LF,i}) \Big)^2 \\ + \frac{1}{N_H} \sum_{i=1}^{N_H} \Big( y_{HF,i}-f_{\theta_{HF},j}\big( f_{\theta_{LF},j}(x_{HF,i}) \big) \Big)^2 \ ,
\label{eq:loss}
\end{multline}
\noindent where $N_L$ and $N_H$ correspond to the low- and high-fidelity batch sizes respectively. Our MF-RPN learns the mappings between related physical variables: emulating the parameterization (inputs to outputs) at low fidelity via the network $f_{\theta_{LF},j}$, and mapping both parameterization outputs at different fidelity levels (low- to high-fidelity) using the network $f_{\theta_{HF},j}$. 

We opted for a joint training of the low- and high-fidelity networks since a sequential training would put more weight and importance on the second network $f_{\theta_{HF},j}$ (blue network in figure \ref{fig:overview}.b) as it will be trained after the fit of network $f_{\theta_{LF},j}$ (red network in figure \ref{fig:overview}.a), which is then held fixed. We observed that a sequential training favors converging to a MF-RPN model that is nearly identical to the SF-HF-RPN model since the last learning step in fitting the MF-RPN model is nearly identical to the SF-HF-RPN model's learning of the mapping between high-fidelity parameterization inputs and outputs.

Another important aspect regarding the chosen architecture of the MF-RPN model is the uncertainty propagation across fidelity levels. Once properly trained, any uncertainty in the paramterization input is propagated to the corresponding low-fidelity parameterization output via the low-fidelity RPN (red network in figure \ref{fig:overview}). These low-fidelity predictions are directly taken as inputs for the second ensemble (blue network in figure \ref{fig:overview}). Hence, their corresponding uncertainty is naturally propagated to the corresponding high-fidelity parameterization output predictions, ensuring a continuous uncertainty propagation from the low- to high-fidelity variables.

Machine learning models usually need to be trained on normalized data. In this work, standardization (or Z-score) is used as normalization so that inputs and outputs have the properties of a Gaussian distribution with a zero mean and unit variance. Since the MF model aggregates two datasets of different fidelity levels and therefore with different distribution supports, a choice regarding the data normalization for the MF model has to be made. On one hand, the MF model is designed to tackle the task of extrapolation beyond the high-fidelity training data. If the latter is chosen for data normalization, then the MF model would be required to make predictions for high-fidelity testing data points, while mapping them with respect to the distribution of the high-fidelity training data corresponding to the SPCAM5 historical run. On the other hand, the low-fidelity training data was built such that it provides the MF model with useful information regarding the extrapolation scenarios. If the MF-RPN model is built based on data normalization using the low-fidelity training data, then its predictions for high-fidelity testing data points will be estimated while mapping testing inputs and outputs with respect to the distribution of the low-fidelity training data corresponding to the CAM5 $+8K$ simulation. This latter scenario of a warmer climate is closer to the high-fidelity extrapolation scenario of interest. As such, the MF-RPN model is trained on data that is normalized with a standardization based on the mean and standard deviation of the low-fidelity data corresponding to the CAM5 $+8K$ run since the extrapolation to a warmer climate is more critical than the data accuracy. In this work, all data are normalized to unit normal distribution. Hence, MF-RPN's inputs and outputs are normalized with respect to the mean and standard deviation of the CAM5 $+8K$ simulation dataset. This normalization applies to the high-fidelity training data: SPCAM5 historical run, and also to the low-fidelity training data: CAM5 $+8K$ run. The chosen normalization helps the MF-RPN model account for the distribution shift between the training and testing high-fidelity data, based only on information of the computationally cheaper but valuable (for extrapolation) low-fidelity training data. Distributions of the normalized test data using statistics from CAM5 $+8K$ and SPCAM5 historical runs confirm the physically-based motivation of using the former dataset for MF-RPN normalization as detailed above (table \ref{tab:stat_test}). Indeed, the normalized test data based on the CAM5 $+8K$ statistics shows variables distributions that are closer to the unit normal one which is the ideal distribution to train the ML-based MF-RPN model on. 

\begin{table*}
\centering
\begin{tabular}{|l|l|l|}
\hline
  & CAM5 $+8K$ statistics & SPCAM5 historical run statistics \\
\hline
Mean of relative humidity & -0.33 & 1.96 \\
\hline
Std. dev. of relative humidity & 0.61 & 2.79 \\
\hline
Mean of heat tendency & 0.01 & -0.006 \\
\hline
Std. dev. of heat tendency & 0.94 & 1.21 \\
\hline
\end{tabular}
\caption{\label{tab:stat_test} Mean and standard deviation of relative humidity and heat tendency for the normalized test data using statistics from CAM5 $+8K$ and SPCAM5 historical runs. Results are averaged across all vertical levels.}
\end{table*}



\subsection*{RPNs' individual networks hyperparamters and training}

Hyperparameters of individual neural networks forming different RPNs models did not need to be tuned from scratch, and were instead chosen based on the hyperparameter optimization over $\sim 250$ trials conducted in Mooers et al.'s study on fully-connected neural network convection superparameterization for SPCAM5 \cite{Mooers2021}. In particular, individual Multi-Layer Perceptrons (MLPs) forming the RPN were considered as fully connected neural networks with $7$ hidden layers, each containing $512$ neurons. We utilized a batch size of $2048$ and ReLU activation (with a negative slope of 0.15) for all layers except for the output one, where the linear activation function was used.

The MLPs were trained for a total of $236 520$ stochastic gradient descent (SGD) steps using the Adam optimizer. The learning rate was initialized at $10^\text{-4}$ with an exponential decay at a rate of $0.99$ per $1000$ steps. For data bootstrapping, each network in the RPN ensembles is trained on a randomly sampled subset with a size equal to $80\%$ of the whole training dataset size as justified in \textit{Yang et al.} \cite{Yang2022}.

\subsection*{Error metrics}

In this section, we define the different error metrics that were used to evaluate the performance of the different surrogate models. We keep the formulation as generic as possible with respect to all paramterization output variables. We also keep the definition general so that it can accommodate the evaluation either on the whole test dataset (points concatenated across space and time) or on a subset of the test dataset (with concatenation along time and/or some specific space dimensions). Global error metrics will be evaluated across all test data points concatenated over space and time. For longitude-latitude structure, the error metrics are evaluated on points concatenated across time. For pressure-latitude structure, the error metrics are evaluated  on points concatenated over time and longitude. Models errors are evaluated on daily averages as performed in \textit{Mooers et al.} \cite{Mooers2021} in order to have a comprehensive assessment of the models performance. For the MAE metric, heat and moisture tendencies are scaled by the specific heat capacity of air at a constant pressure ($1004.6 \mathrm{J.kg^{-1}.K^{-1}}$) and latent heat of vaporization at standard atmospheric conditions ($2.26\times10^6 \mathrm{J.kg^{-1}}$), respectively \cite{Mooers2021}. In the next section $y$ denotes the true target value and $\hat{y}$ the corresponding prediction. $\mathcal{D}$ will denote the test dataset and $| \mathcal{D} |$ its size.


\subsubsection*{Mean Absolute Error (MAE):}
\begin{equation}
\text{MAE} = \frac{1}{ |\mathcal{D}| }\sum_{i\in \mathcal{D}}|y_i - \hat{y}_i| \ ,
\end{equation}
\noindent where $\mathcal{D}$ denotes the test dataset, $y_i$ the true target and  $\hat{y}_i$ the corresponding model prediction.

For global error evaluation, $\mathcal{D}$ corresponds to the whole test dataset (points concatenated across space and time). For the longitude-latitude plots, $\mathcal{D}$ corresponds to the test dataset concatenated across time, providing a single error metric evaluation for each parameterization output variable and for each point in longitude-latitude cross-section. 

For pressure-latitude plots, $\mathcal{D}$ corresponds to the test dataset concatenated across time and longitude dimension. Hence, for these plots, each pressure level corresponds to a specific paramterization output variable, and each point in latitude has a single error metric evaluation for each pressure level. For the temporal error evaluation, $\mathcal{D}$ corresponds to the test dataset concatenated across longitude and latitude dimensions.

\subsubsection*{Coefficient of Determination ($R^2$):}

\begin{equation}
\text{R$^\text{2}$} = 1 - \frac{\sum_{i\in \mathcal{D}}(y_i - \hat{y}_i)^2}{\sum_{i\in \mathcal{D}}(y_i - \bar{y})^2}
\end{equation}

where $\bar{y}$ represents the true target value averaged over the test dataset $\mathcal{D}$. The definition of the different choices for the test dataset $\mathcal{D}$ is the same as detailed above for MAE.

\subsubsection*{Stochastic Metric (CRPS):}

The Continuous Ranked Probability Score (CRPS) \cite{James1976,Gneiting2007} is a generalization of the MAE for distributional predictions. CRPS penalizes over-confidence in addition to inaccuracy in ensemble predictions. A lower CRPS value corresponds to a more accurate and/or less over-confident model. For each variable, it measures the discrepancy between the ground truth target $y$ with the cumulative distribution function (CDF) $\hat{F}$ of the prediction via:

\begin{multline}
\mathrm{CRPS}(\hat{F}, y) = \int \left( \hat{F}(z) - \mathbf{1}_{\{z \geq y\}} \right)^2 dz \\
= \mathbb{E}[|\hat{y} - y|] - \frac{1}{2} \mathbb{E}[|\hat{y} - \hat{y}'|]
\end{multline}

where $\hat{y}, \hat{y}' \sim \hat{F}$ are independent and identically distributed ($iid$) samples from the predicted CDF. We use the following non-parametric estimate form of the CRPS \cite{Ferro2014}:

\begin{equation}
\mathrm{CRPS}(\mathbf{\hat{y}}, y) = \frac{1}{n} \sum_{i=1}^n |\hat{y}_i - y| - \frac{1}{2 n (n-1)} \sum_{i=1}^n \sum_{j=1}^n |\hat{y}_i - \hat{y}_j|, \label{crps}
\end{equation}

where the CDF $\hat{F}$ is estimated empirically using $n = 32\;iid$ samples $\hat{y}_i \sim \hat{F}$. Equation \eqref{crps} corresponds to the CRPS estimate for a singular datapoint. For a given test dataset $\mathcal{D}$, the corresponding CRPS is obtained as an average of individual CRPS estimates \eqref{crps} over all datapoints within $\mathcal{D}$. The first term in \eqref{crps} is the MAE between the target and samples of the predictive distribution, while the second term is smaller for smaller predictive variances and vanishes completely for point estimates. The CRPS definition is naturally extended to the ensemble models by taking each ensemble member prediction as a sample of an implicit predictive distribution. 

\bibliography{main.bib}

\section*{Acknowledgements}

The authors would like to acknowledge funding from the National Science Foundation LEAP Science and Technology center Award \# 2019625, USMILE European Research Council (ERC) Synergy grant, National Science Foundation funding from an AGS-PRF Fellowship Award (AGS2218197) and Department of Energy.

\section*{Author contributions statement}

M.A.B. and P.G. conceived the experiment(s), L.P. generated the training data with guidance by M.P., M.A.B. conducted the experiment(s),  M.A.B. and P.G. analyzed the results,  M.A.B., P.G. and M.P. provided funding, M.A.B., P.G., L.P. and M.P. wrote the manuscript. 

\section*{Additional information}

Following paper publication, all code and data accompanying this manuscript will be made publicly available at \url{https://github.com/bhouri0412/MF_RPN_SPCAM_cv_param}. 

Authors declare that they have no known competing financial interests or personal relationships that could have appeared to influence the work reported in this paper.



\clearpage

\onecolumn

\section*{Supplementary Information}

\subsection*{Global errors}

\begin{figure*}[ht]
\centering
\includegraphics[width=0.8\linewidth]{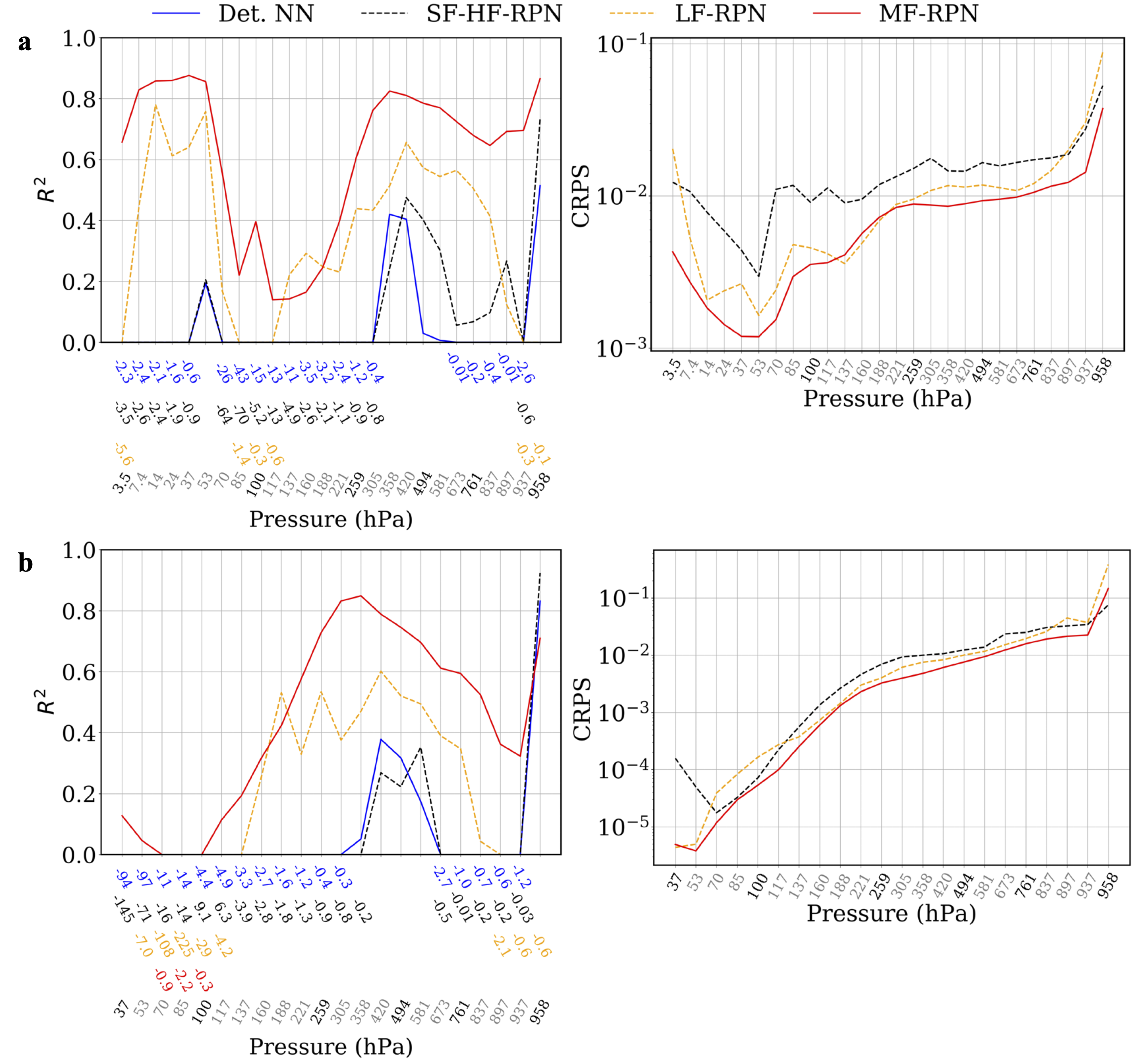}
\caption{\textbf{$\mathbf{R^2}$ and CRPS for different models across all test data points concatenated over space and time}. Y-axes of $R^2$ plots are limited to $[0,1]$ for clarity purposes. Negative $R^2$ values are indicated below the plot with a $60^\circ$ clockwise rotation and the color corresponding to the one used to plot the specific model's results. \textbf{a}, Heat tendency results. \textbf{b}, Moisture tendency results}
\label{fig:glob_R_CRPS_grey_with_negative}
\end{figure*}

For the heat tendency and the vertical levels where the deterministic NN shows significantly negative $R^2$ values, the SF-HF-RPN $R^2$ values are also negative but very similar to those of the deterministic NN (between $3.5$ and $305 \ \mathrm{hPa}$, figure \ref{fig:glob_R_CRPS_grey_with_negative}.a). However, for the vertical levels where the deterministic NN shows nearly zero to slightly negative $R^2$ values (from $494$ to $897 \ \mathrm{hPa}$), the SF-HF-RPN does significantly improve upon the deterministic NN, showing the ability of the RPN surrogate model to resolve part of the convection stochasticity. 

For the moisture tendency and as mentioned in the Results section, the negative $R^2$ values for the deterministic NN. and SF-HF-RPN show that the latter resolves better the moisture convection stochasticity for vertical levels below $494 \ \mathrm{hPa}$, while it does not improve upon the deterministic NN. for higher levels (figure \ref{fig:glob_R_CRPS_grey_with_negative}.b).

The CRPS curves follow a similar trend as the MAE ones (see Results), showing that all RPN-based models have similar confidence in terms of variance prediction (figure \ref{fig:glob_R_CRPS_grey_with_negative}). This behavior could be expected since all these models are based on the same numerical ensemble technique of RPN. Nonetheless, the uncertainty quantification returned by these different RPN-based models are quite different and do not vary similarly as a function of the actual error (see Uncertainty quantification).

\subsection*{Longitude-latitude errors structure}

Figure \ref{fig:long_lat_moist_all} shows the longitude-latitude structure of the $R^2$ and MAE metrics for moisture tendency at vertical levels $259$, $494$ and $761 \ \text{hPa}$ for all surrogate models and evaluated on the test dataset. For the $R^2$ metric, the moisture tendency results for all pressure levels are quite similar with the SF-HF-RPN model improving upon the deterministic NN model in the temperate zone, and the LF-RPN model improving further upon the SF-HF-RPN model within the temperate zone and even within the tropics and polar regions. 

For all vertical levels considered, the MF-RPN model improves even further compared to all other models across all regions, while still showing a few negative $R^2$ values for this extrapolation task like all other models. For the pressure level $494 \ \text{hPa}$, the negative $R^2$ regions for the MF-RPN model include part of the South-East Pacific region, of the south Atlantic ocean and of the African Sahara. However, all these regions are significantly smaller for the pressure level $259 \ \text{hPa}$ compared to the regions observed for $494 \ \text{hPa}$ (white regions in MF-RPN plots in figures \ref{fig:long_lat_moist_all}.a and   \ref{fig:long_lat_moist_all}.b). Similarly the negative $R^2$ regions for the MF-RPN model within the South-East Pacific region and south Atlantic ocean are clearly smaller at the pressure level $761 \ \text{hPa}$ compared to level $494 \ \text{hPa}$ (white regions in MF-RPN plots in figures \ref{fig:long_lat_moist_all}.b and  \ref{fig:long_lat_moist_all}.c). Hence, the MF-RPN model is capable of better extrapolating at relatively low- and high-altitude levels within the troposphere ($259$ and $761 \ \text{hPa}$) compared to mid-altitude levels (around $494 \ \text{hPa}$).

Since the highest error is observed within the tropical regions, the MAE longitude-latitude plots (figures \ref{fig:long_lat_moist_all}.d - \ref{fig:long_lat_moist_all}.f) give a clear representation of the performance improvement within these regions across different models. For all pressure levels considered, the LF-RPN systematically improves the moisture convection parameterization within the tropics compared to the deterministic NN. and SF-HF-RPN models, while the MF-RPN improves even further upon the LF-RPN model. Comparing the deterministic NN. and SF-HF-RPN results, the latter is capable of better resolving the moisture convection stochasticity within the tropics for low vertical levels ($761 \ \text{hPa}$ as seen in figure \ref{fig:long_lat_moist_all}.f), while it fails to do so for higher levels ($259$ and $494 \ \text{hPa}$ as seen in figures \ref{fig:long_lat_moist_all}.d and \ref{fig:long_lat_moist_all}.e).

\begin{figure*}[ht]
\centering
\includegraphics[width=\linewidth]{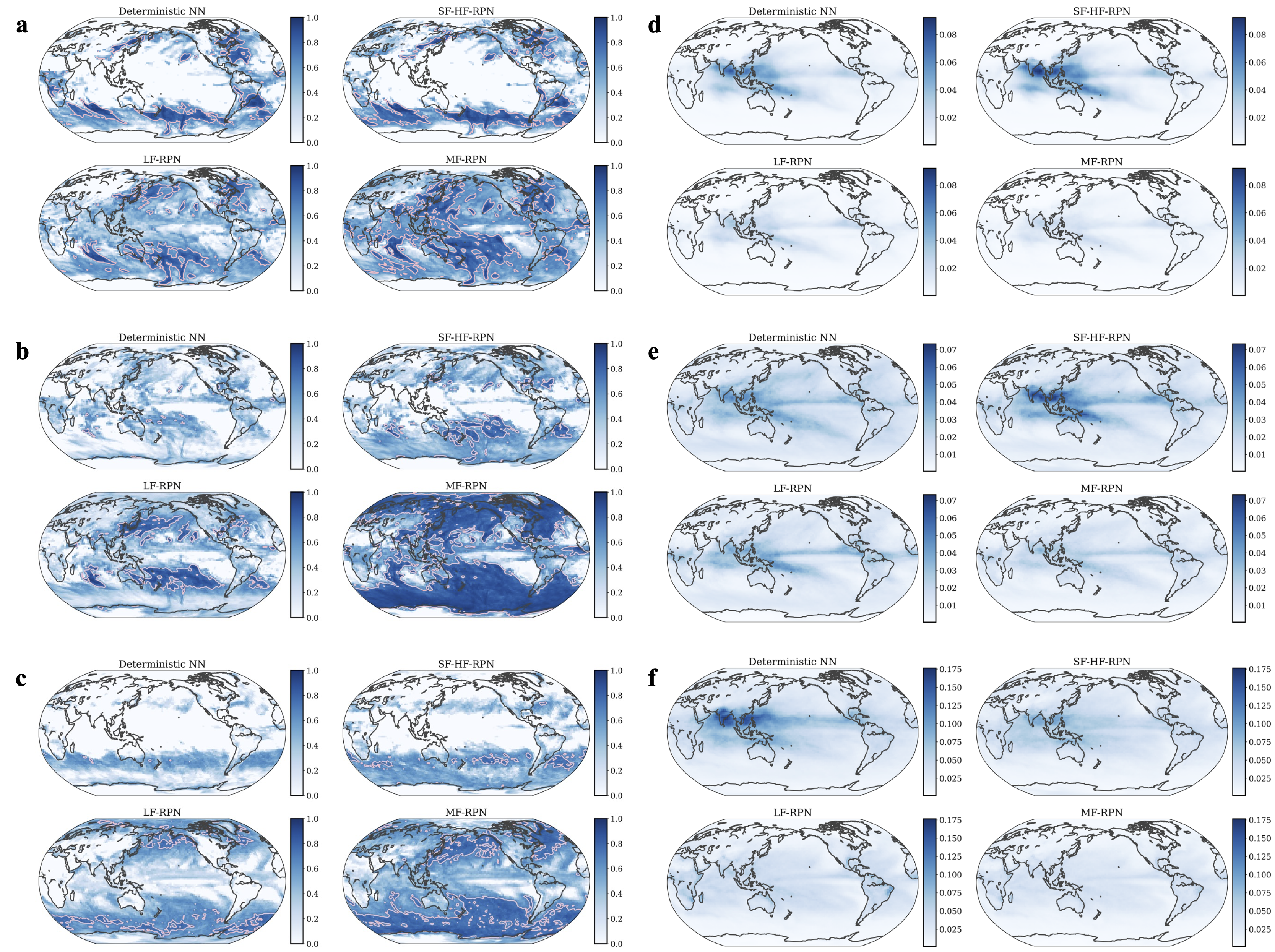}
\caption{\textbf{Longitude-latitude variation of MAE and $\mathbf{R^2}$ for moisture tendency at different vertical levels}. $\mathbf{R^2}$ is evaluated on the test dataset and negative values are lumped to $0$ for clarity purposes. \textbf{a}, $R^2$ at $P = 259 \ \text{hPa}$. \textbf{b}, $R^2$ at $P = 494 \ \text{hPa}$. \textbf{c}, $R^2$ at  $P = 761 \ \text{hPa}$. \textbf{d}, MAE at $P = 259 \ \text{hPa}$. \textbf{e}, MAE at $P = 494 \ \text{hPa}$. \textbf{f}, MAE at  $P = 761 \ \text{hPa}$. It is an encouraging sign of successful stochastic learning that RPN predictions are most uncertain in the same tropical locations where determinstic errors are highest.}
\label{fig:long_lat_moist_all}
\end{figure*}

\begin{figure*}[ht]
\centering
\includegraphics[width=\linewidth]{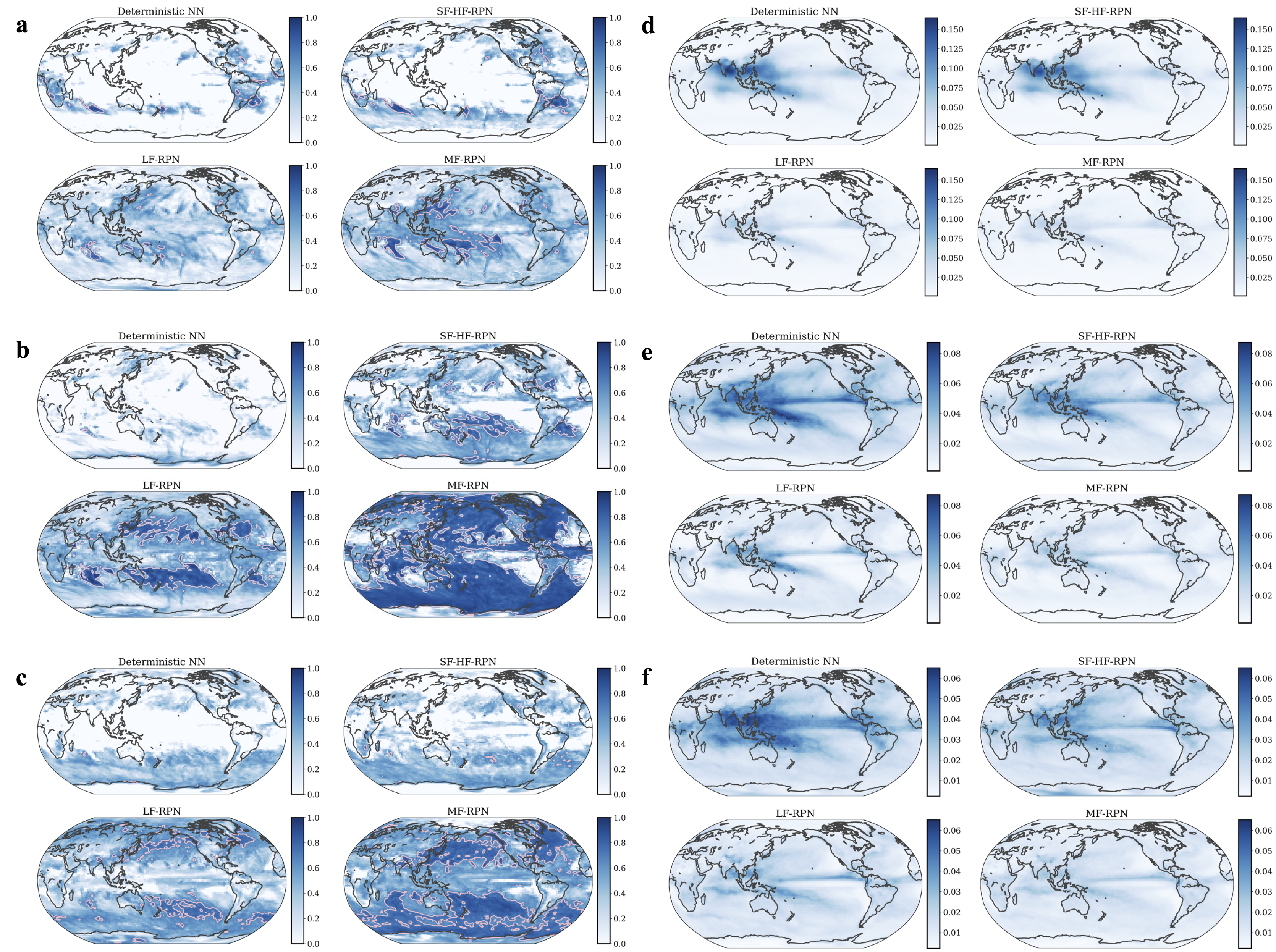}
\caption{\textbf{Longitude-latitude variation of MAE and $\mathbf{R^2}$ for heat tendency at different vertical levels}. $\mathbf{R^2}$ is evaluated on the test dataset and negative values are lumped to $0$ for clarity purposes. \textbf{a}, $R^2$ at $P = 259 \ \text{hPa}$. \textbf{b}, $R^2$ at $P = 494 \ \text{hPa}$. \textbf{c}, $R^2$ at  $P = 761 \ \text{hPa}$. \textbf{d}, MAE at $P = 259 \ \text{hPa}$. \textbf{e}, MAE at $P = 494 \ \text{hPa}$. \textbf{f}, MAE at  $P = 761 \ \text{hPa}$.}
\label{fig:long_lat_heat_all}
\end{figure*}

Figure \ref{fig:long_lat_heat_all} shows the longitude-latitude structure of the $R^2$ and MAE metrics for heat tendency at vertical levels $259$, $494$ and $761 \ \text{hPa}$ for all surrogate models and evaluated on the test dataset. For the $R^2$ variation, the heat tendency results for all pressure levels are quite similar to the results obtained for the moisture tendency, with the SF-HF-RPN model improving upon the deterministic NN. one in the temperate zone, and the LF-RPN model improving further upon the SF-HF-RPN model within the temperate zone and even within the tropics and polar regions. 

For all vertical levels considered, the MF-RPN model improves even further compared to all other models across all regions, while still showing some negative $R^2$ values for this extrapolation task like all other models. For the pressure level $494 \ \text{hPa}$, the negative $R^2$ regions for the MF-RPN model include part of the South-East Pacific region, of the south Atlantic ocean and of the African Sahara as observed for the moisture tendency. However, all these regions are significantly smaller and nearly non-existent for the pressure levels $259$ and $761 \ \text{hPa}$ compared to the regions observed for $494 \ \text{hPa}$ (white regions in  MF-RPN plots in figures \ref{fig:long_lat_heat_all}.a-\ref{fig:long_lat_heat_all}.c). Hence, the MF-RPN model is capable of better extrapolating at relatively low and high altitude levels within the troposphere ($259$ and $761 \ \text{hPa}$) compared to mid altitude levels (around $494 \ \text{hPa}$) as observed for the moisture tendency. 

It is worth noting that at vertical level $494 \ \text{hPa}$, the LF-RPN model shows better results compared to the MF-RPN model in the South-East Pacific region, south Atlantic ocean and the African Sahara (figure \ref{fig:long_lat_heat_all}.b). In these regions, the LF-RPN model returns overall positive $R^2$ values, while the MF-RPN fails to do so. This result suggests that the LF-RPN model is better at approximating shallow convection, while the MF-RPN model does a better job at learning deep convection. The difference in behavior can be attributed to the different datasets on which these two models have been trained since the LF-RPN is only trained on the 1-year CAM5 $+8K$ simulation, while the MF-RPN model has to aggregate both the latter dataset with the 3-month SPCAM5 historical run. Nonetheless, at the vertical level of $494 \ \text{hPa}$, the MF-RPN clearly outperforms the LF-RPN model in all other tropical regions except the three ones listed above showing again the overall better generalization and extrapolation achieved by the MF model.

The MAE longitude-latitude plots (figures \ref{fig:long_lat_heat_all}.d - \ref{fig:long_lat_heat_all}.f) clearly show the improvement obtained for the heat convection parameterization with the MF-RPN compared to the LF-RPN for all pressure levels considered. The LF-RPN model also improves upon the deterministic NN. and SF-HF-RPN models. However, unlike the moisture tendency results, the SF-HF-RPN model does not show higher errors compared to the deterministic NN. model at pressure level $259 \ \text{hPa}$ (figure \ref{fig:long_lat_heat_all}.d), and does improve the heat convection parameterization at pressure levels  $494$ and $761 \ \text{hPa}$ (figures \ref{fig:long_lat_heat_all}.e and \ref{fig:long_lat_heat_all}.f). Hence, the SF-HF-RPN is capable of better resolving the  convection stochasticity across different vertical levels for the heat tendency compared to the moisture one.

\subsection*{Pressure-latitude errors structure}

\begin{figure*}
\centering
\includegraphics[width=0.9\linewidth]{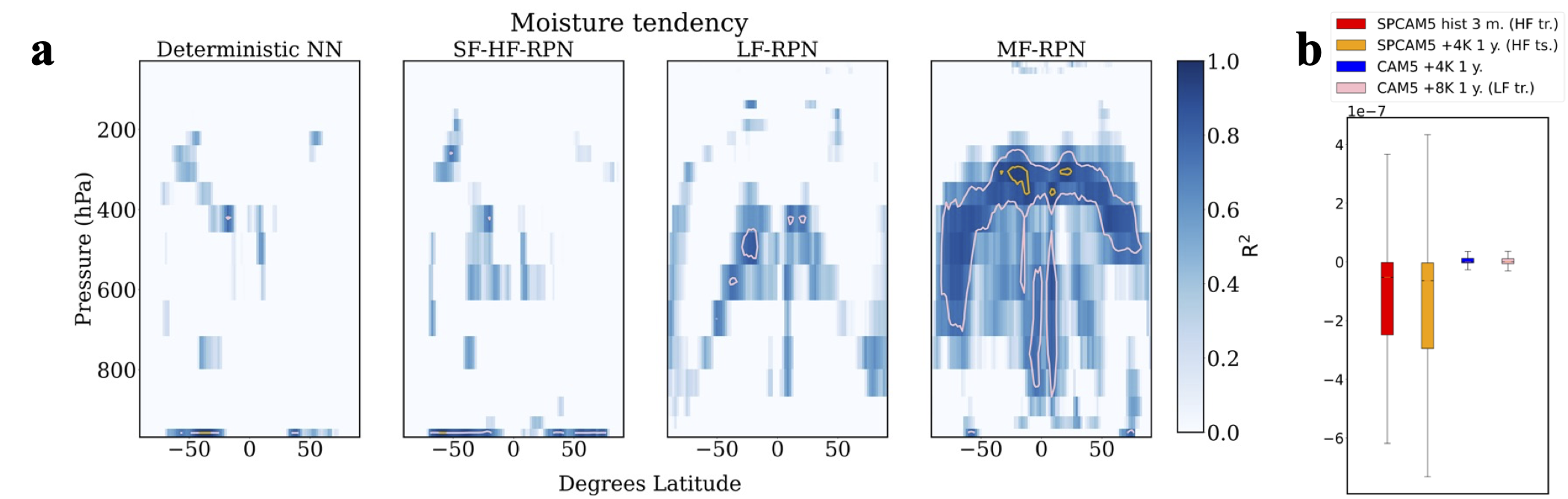}
\caption{\textbf{Pressure-latitude variation of coefficient of determination $\mathbf{R^2}$ for moisture tendency and different surrogate models}. $\mathbf{R^2}$ is evaluated on the test dataset and negative values are lumped to $0$ for clarity purposes. \textbf{a}, Moisture tendency results. \textbf{b}, Data distribution of moisture tendency for different datasets at lowest vertical level at $958 \ hPa$.}
\label{fig:press_lat_heat_moist_rgriff_w_data_dist}
\end{figure*}

Based on the data distribution of moisture tendency for the lowest vertical level at $958 \ \mathrm{hPa}$ (figure \ref{fig:press_lat_heat_moist_rgriff_w_data_dist}.b), it is quite obvious that for this particular level, the low-fidelity training data of CAM5 $+8K$ simulation is not informative on the extrapolation scenario of interest when compared to data distributions of HF training and testing data. Similarly, the CAM5 $+4K$ is not informative neither as the corresponding data spans a similar interval to the CAM5 $+8K$ dataset. For other vertical levels, the CAM5 $+8K$ simulation dataset does extrapolate beyond the HF training data (see Methods). These datasets distributions explain the overall lower performance observed for LF-RPN and MF-RPN models when it comes to infer the moisture tendency at the lowest vertical level $958 \ \mathrm{hPa}$ (figure \ref{fig:press_lat_heat_moist_rgriff_w_data_dist}.a).

\subsection*{Temporal errors structure}

\begin{figure*}[ht]
\centering
\includegraphics[width=\linewidth]{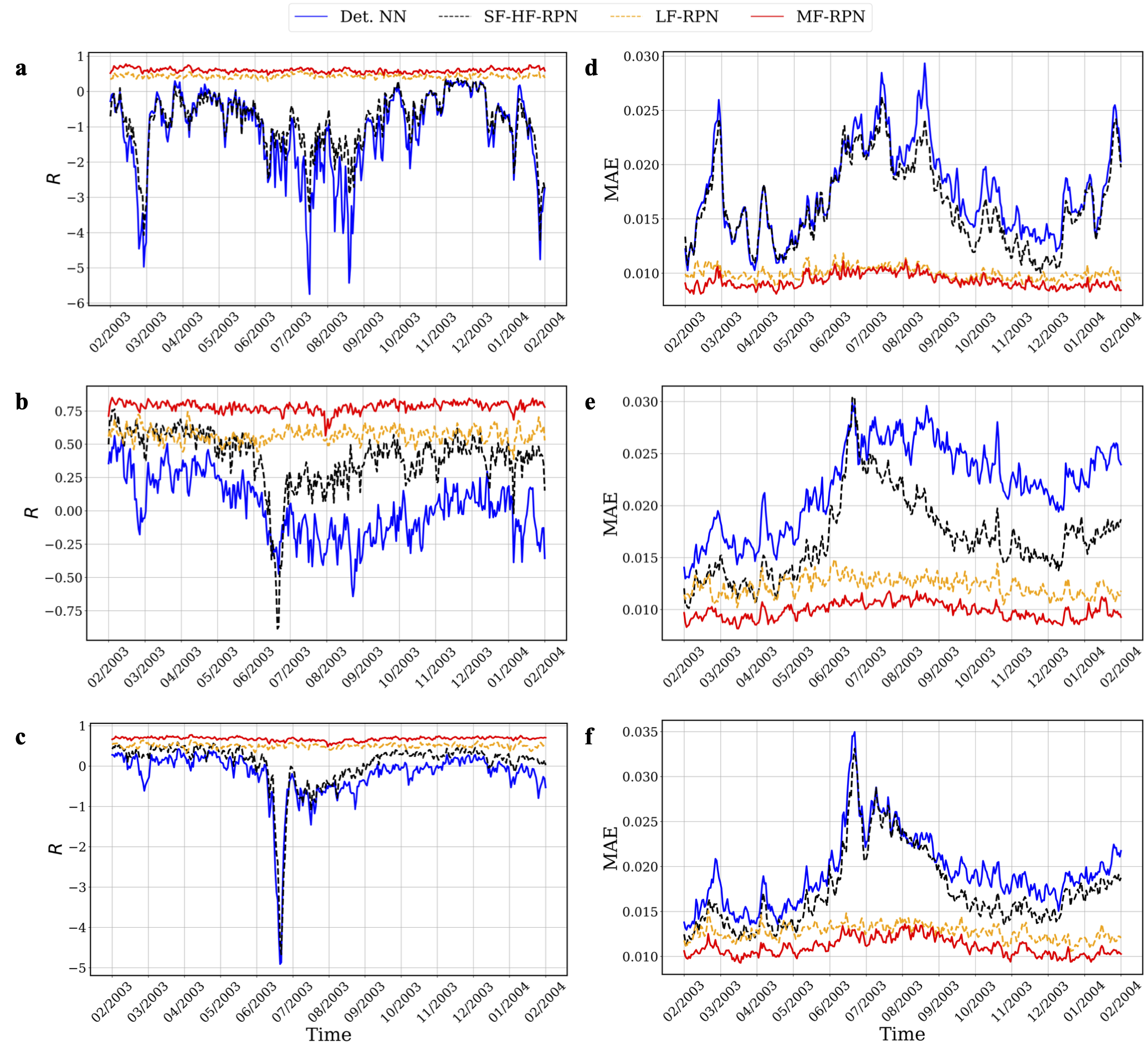}
\caption{\textbf{Temporal variation of MAE and $\mathbf{R^2}$ for heat tendency at different vertical levels and evaluated on test dataset}. \textbf{a}, $R^2$ at $P = 259 \ \text{hPa}$. \textbf{b}, $R^2$ at $P = 494 \ \text{hPa}$. \textbf{c}, $R^2$ at  $P = 761 \ \text{hPa}$. \textbf{d}, MAE at $P = 259 \ \text{hPa}$. \textbf{e}, MAE at $P = 494 \ \text{hPa}$. \textbf{f}, MAE at  $P = 761 \ \text{hPa}$.}
\label{fig:temp_heat_all_120}
\end{figure*}

Figure \ref{fig:temp_heat_all_120} shows the temporal variations of testing MAE and $\mathbf{R^2}$ for heat tendency and different surrogate models at different vertical levels. In coherence with the global errors and spatial structures of different error metrics, the MF-RPN is the best performing model across all testing period and different vertical levels, followed by the LF-RPN model. The errors temporal variations show a better performance overall for the SF-HF-RPN model compared to the deterministic NN. mostly for the vertical level $494 \ \text{hPa}$, while the improvement for the two other vertical levels is mainly limited to the second half of the testing period of a year. 

The seasonality effect on the models performance is not quite trivial. We remind that the SF-HF-RPN and deterministic NN. are trained on an SPCAM5 historical run simulation from February 1st 2003 to April 31st 2003, while the LF-RPN is trained (jointly to the MF-RPN training) on a CAM5 $+8K$ simulation from February 1st 2003 to January 31st 2004. The MF-RPN model is trained on both datasets. The test dataset corresponds to an SPCAM5 $+4K$ run simulation from February 1st 2003 to January 31st 2004. However, the temporal variations of the SF-HF-RPN and deterministic NN. testing errors show poor performance even in the period between February 1st 2003 and April 31st 2003 (figures \ref{fig:temp_heat_all_120}.a and \ref{fig:temp_heat_all_120}.d). Hence, the SF-HF-RPN and deterministic NN.'s error temporal variations for the vertical level $259 \ \text{hPa}$ are mostly governed by extrapolating to a different climate scenario rather than extrapolating to a different seasonality. On the other hand, their error temporal variations for the vertical level $494 \ \text{hPa}$ do show a significantly lower performance when extrapolating beyond the training period (figures \ref{fig:temp_heat_all_120}.b and \ref{fig:temp_heat_all_120}.e). Therefore, the testing error for the vertical level $494 \ \text{hPa}$ is affected not only by the extrapolation to a different climate but also to a different seasonality. The SF-HF-RPN and deterministic NN.'s error temporal variations for the vertical level $761 \ \text{hPa}$ show lower performance around the period between June and August, but the values observed for the second half of the testing year are similar to those obtained for the period between February and May (figures \ref{fig:temp_heat_all_120}.c and \ref{fig:temp_heat_all_120}.f). Hence, the temporal extrapolation for the vertical level $761 \ \text{hPa}$ seems to affect the error differently based on the unseen season among the training dataset. Note that since the testing dataset corresponds to a climate simulation that was not considered in any of the training datasets, it is always challenging to disentangle the different extrapolations contribution on the models errors. Given the heterogeneous temporal variations of the SF-HF-RPN and deterministic NN. errors for different seasons across different vertical levels, it is not straightforward to draw conclusions on the seasonality effect on these models performance.

\begin{figure*}[ht]
\centering
\includegraphics[width=\linewidth]{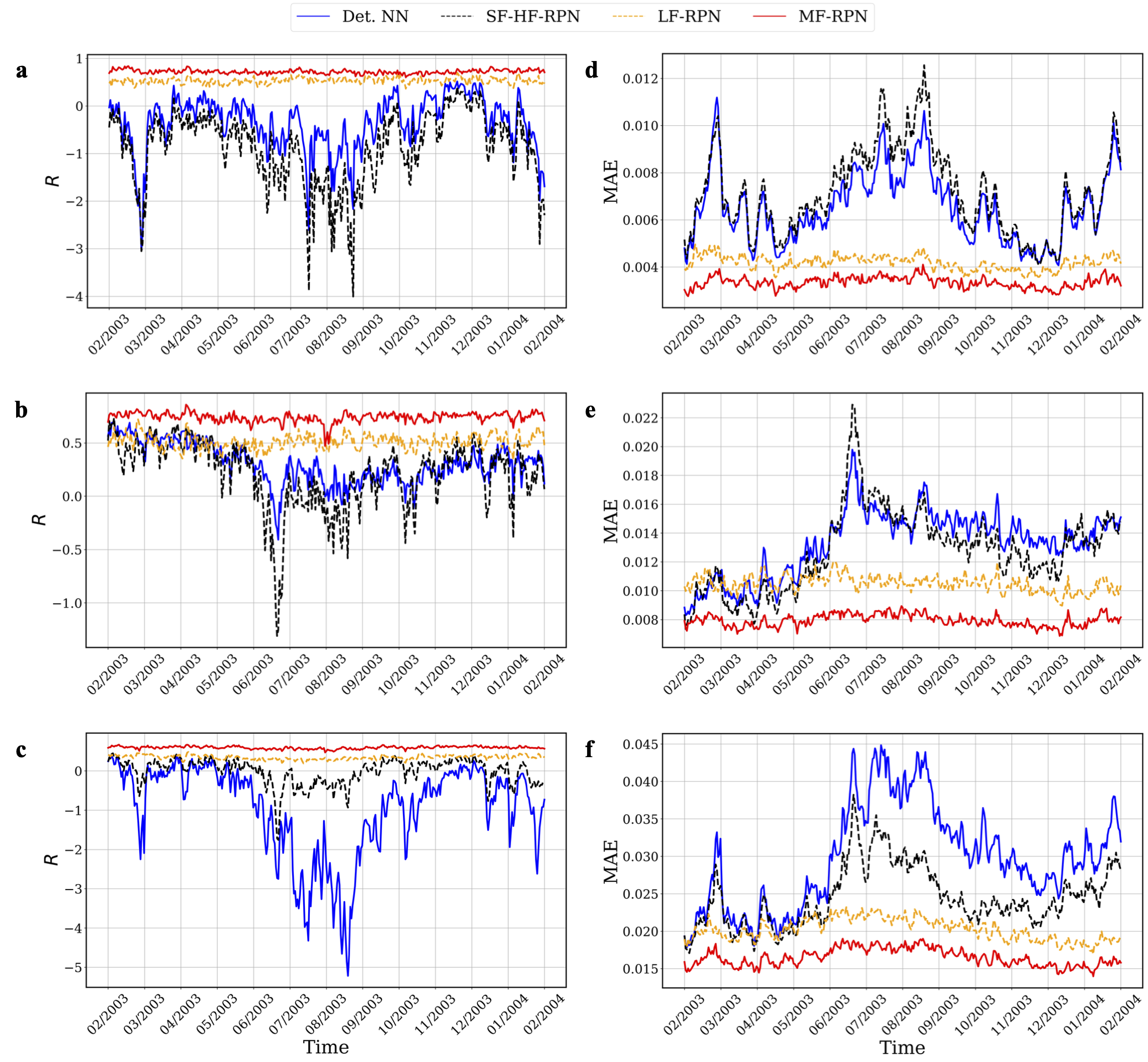}
\caption{\textbf{Temporal variation of MAE and $\mathbf{R^2}$ for moisture tendency at different vertical levels and evaluated on test dataset}. \textbf{a}, $R^2$ at $P = 259 \ \text{hPa}$. \textbf{b}, $R^2$ at $P = 494 \ \text{hPa}$. \textbf{c}, $R^2$ at  $P = 761 \ \text{hPa}$. \textbf{d}, MAE at $P = 259 \ \text{hPa}$. \textbf{e}, MAE at $P = 494 \ \text{hPa}$. \textbf{f}, MAE at  $P = 761 \ \text{hPa}$.}
\label{fig:temp_moist_all_120}
\end{figure*}

Figure \ref{fig:temp_moist_all_120} shows the temporal variations of testing MAE and $\mathbf{R^2}$ for moisture tendency and different surrogate models at different vertical levels. In coherence with the global errors and spatial structures of different error metrics, the MF-RPN is the best performing model across all testing period and different vertical levels, followed by the LF-RPN model. The errors temporal variations show a better performance for the SF-HF-RPN model compared to the deterministic NN. only for the vertical level $761 \ \text{hPa}$, while the deterministic NN. outperforms the SF-HF-RPN model for vertical levels $259$ and $494 \ \text{hPa}$. These results are well coherent with the global errors and the longitude-latitude structures of $R^2$, confirming that the SF-HF-RPN is more capable of resolving the heat convection stochasticity and is only able of better resolving the moisture convection stochasticity for vertical levels below the $494 \mathrm{hPa}$ one, while it struggles to do so for higher levels.

As observed for the heat convection parameterization, the seasonality effect on the models performance is not quite trivial. Indeed, for the vertical level $259 \ \text{hPa}$ the temporal variations of the SF-HF-RPN and deterministic NN. testing errors show poor performance even in the period between February 1st 2003 and April 31st 2003 (figures \ref{fig:temp_moist_all_120}.a and \ref{fig:temp_moist_all_120}.d). On the other hand, their error temporal variations for the vertical levels $494$ and $761 \ \text{hPa}$ do show a significantly lower performance when extrapolating beyond the training period (figures \ref{fig:temp_moist_all_120}.b, \ref{fig:temp_moist_all_120}.c, \ref{fig:temp_moist_all_120}.e and \ref{fig:temp_moist_all_120}.f). These results along with those obtained for the heat convection parameterization confirm that it is not straightforward to draw conclusions on the seasonality effect on these models performance.

\subsection*{Uncertainty quantification}

In addition to the deterministic and statistical error metrics, we compare the performance of different Bayesian surrogate models with respect to the returned uncertainty quantification. For a given input $x$, each of the RPN-based models returns an ensemble of predictions from which we infer the mean $\hat{y}(x)$ for each of the output variables. The mean is used to estimate the different error metrics discussed above. We can also estimate the corresponding standard deviation $\sigma_M(x)$ which serves as an uncertainty quantification. We use the subscript $M$ to emphasize that we consider the model's inherent uncertainty that is estimated for each individual input point. Standard deterministic machine learning-based parameterizations do not provide the inherent uncertainty and can only return uncertainties by averaging over several input points (e.g. different input points across space and/or time which gives uncertainties that are ``contaminated" with spatial and/or temporal physical variations).

\begin{figure*}[ht]
\centering
\includegraphics[width=0.65\linewidth]{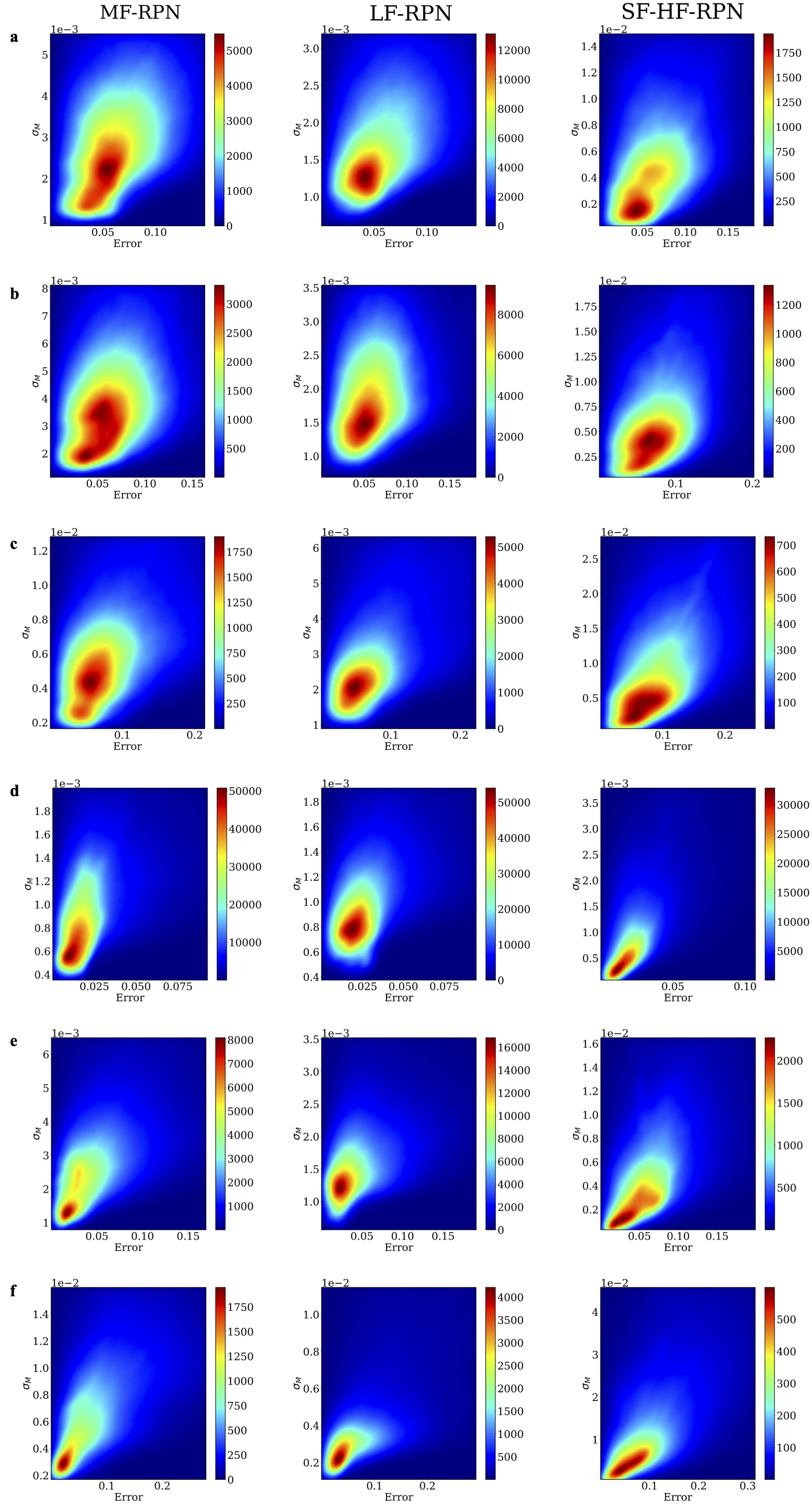}
\caption{\textbf{Density plot of uncertainty $\mathbf{\sigma_M}$ as a function of error for testing points concatenated over space and time for MF-RPN, LF-RPN and SF-HF-RPN, heat and moisture tendencies and at different vertical levels}. \textbf{a}, Heat tendency at $P = 259 \ \text{hPa}$. \textbf{b}, Heat tendency at $P = 494 \ \text{hPa}$. \textbf{c}, Heat tendency at  $P = 761 \ \text{hPa}$. \textbf{d}, Moisture tendency at $P = 259 \ \text{hPa}$. \textbf{e}, Moisture tendency at $P = 494 \ \text{hPa}$. \textbf{f}, Moisture tendency at  $P = 761 \ \text{hPa}$. }
\label{fig:uncert_density_all_models}
\end{figure*}

Figure \ref{fig:uncert_density_all_models} shows the density plots of the uncertainty $\sigma_M$ as a function of the prediction error evaluated on the test dataset for different RPN-based models. It includes the results for heat and moisture tendencies at vertical levels $259$, $494$ and $761 \ \text{hPa}$. A perfect uncertainty quantification would strictly increase with the model's error. For the heat tendency, the MF-RPN's uncertainty clearly displays a more increasing behavior with the error compared to the SF-HF-RPN and LF-RPN models for all vertical levels since the red and yellow regions stretch over larger areas indicating a wider increase of the uncertainty as the error grows (figures \ref{fig:uncert_density_all_models}.a - \ref{fig:uncert_density_all_models}.c). For instance, for the vertical level $P = 259 \ \text{hPa}$, the LF-RPN and SF-HF-RPN density plots clearly show narrow red regions where the returned uncertainty does not vary much compared to the error, unlike the MF-RPN's estimated uncertainty (figure \ref{fig:uncert_density_all_models}.a).

For the moisture tendency, the MF-RPN's uncertainty still displays a more increasing behavior with the error compared to the SF-HF-RPN and LF-RPN models for all vertical levels since the yellow regions stretch over larger areas indicating a wider increase of the uncertainty as the error grows (figures \ref{fig:uncert_density_all_models}.d - \ref{fig:uncert_density_all_models}.f). However, it is worth noting that the SF-HF-RPN's uncertainty density plot shows wider red regions for vertical levels $494$ and $761 \ \text{hPa}$ compared to the MF-RPN (figures \ref{fig:uncert_density_all_models}.e and \ref{fig:uncert_density_all_models}.f). Hence, the SF-HF-RPN's uncertainty displays a more increasing behavior with the error for low error values. Nonetheless, the density plots yellow regions are larger for the MF-RPN model compared to the SF-HF-RPN, highlighting that the MF-RPN's uncertainty is more accurate for moderate to high error values.

\begin{figure*}[ht]
\centering
\includegraphics[width=\linewidth]{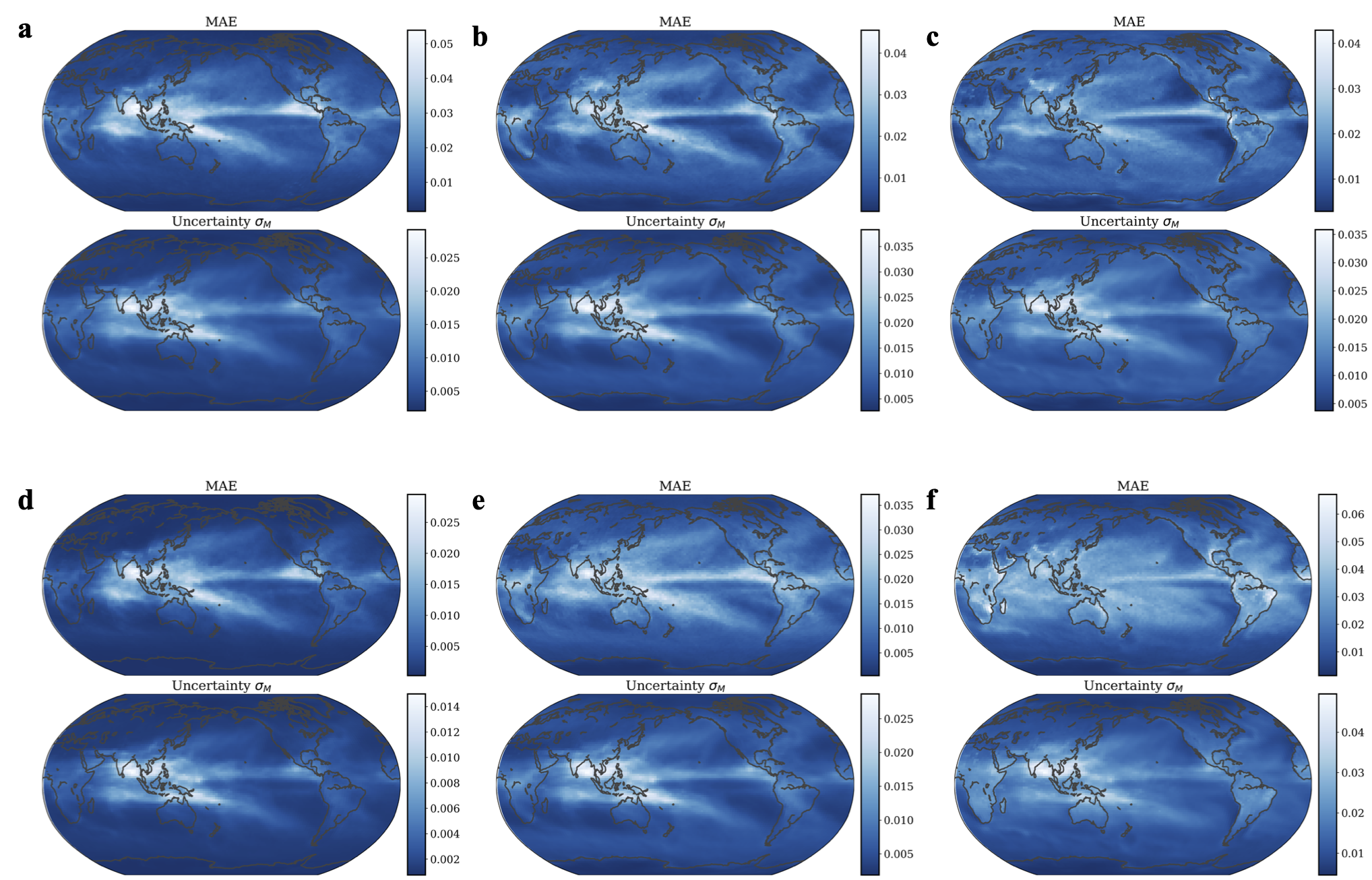}
\caption{\textbf{Longitude-latitude variation of testing MAE and quantified uncertainty $\mathbf{\sigma_M}$ for MF-RPN for heat and moisture tendencies at different vertical levels}. \textbf{a}, Heat tendency at $P = 259 \ \text{hPa}$. \textbf{b}, Heat tendency at $P = 494 \ \text{hPa}$. \textbf{c}, Heat tendency at  $P = 761 \ \text{hPa}$. \textbf{d}, Moisture tendency at $P = 259 \ \text{hPa}$. \textbf{e}, Moisture tendency at $P = 494 \ \text{hPa}$. \textbf{f}, Moisture tendency at  $P = 761 \ \text{hPa}$.}
\label{fig:uncert_long_lat_all}
\end{figure*}

\begin{figure*}[ht]
\centering
\includegraphics[width=\linewidth]{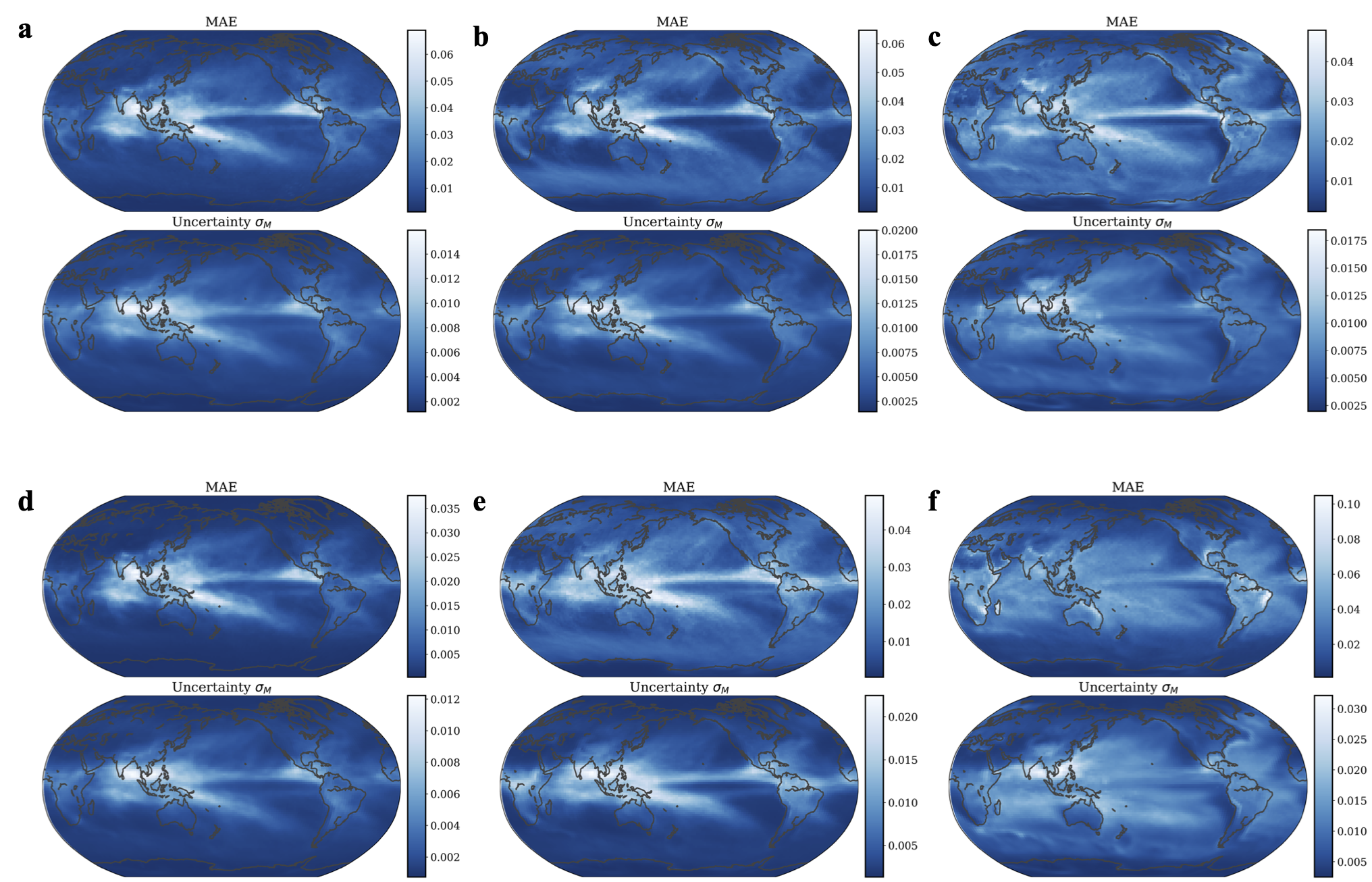}
\caption{\textbf{Longitude-latitude variation of testing MAE and quantified uncertainty $\mathbf{\sigma_M}$ for LF-RPN for heat and moisture tendencies at different vertical levels}. \textbf{a}, Heat tendency at $P = 259 \ \text{hPa}$. \textbf{b}, Heat tendency at $P = 494 \ \text{hPa}$. \textbf{c}, Heat tendency at  $P = 761 \ \text{hPa}$. \textbf{d}, Moisture tendency at $P = 259 \ \text{hPa}$. \textbf{e}, Moisture tendency at $P = 494 \ \text{hPa}$. \textbf{f}, Moisture tendency at  $P = 761 \ \text{hPa}$.}
\label{fig:uncert_long_lat_all_LF}
\end{figure*}

\begin{figure*}[ht]
\centering
\includegraphics[width=\linewidth]{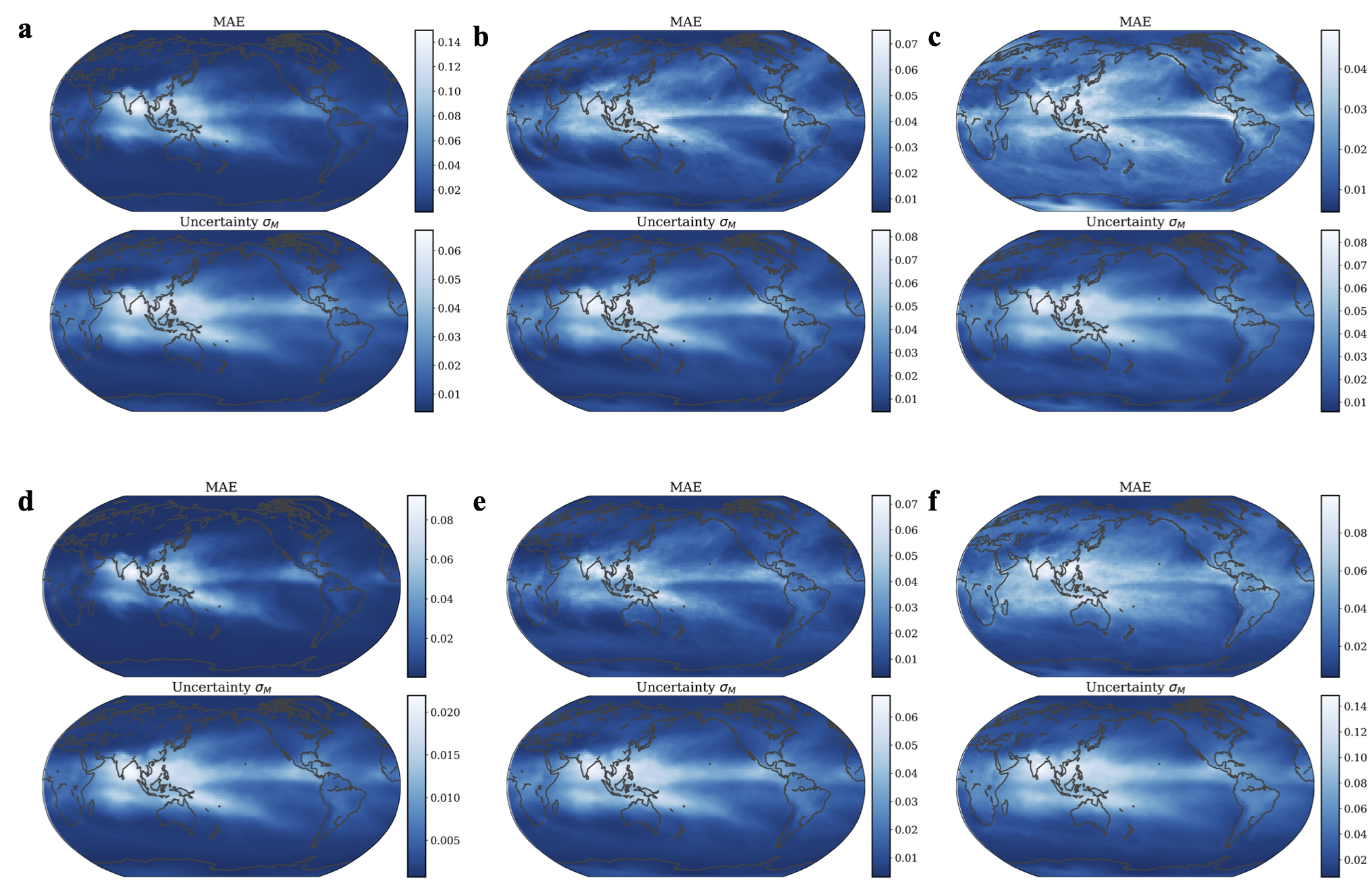}
\caption{\textbf{Longitude-latitude variation of testing MAE and quantified uncertainty $\mathbf{\sigma_M}$ for SF-HF-RPN for heat and moisture tendencies at different vertical levels}. \textbf{a}, Heat tendency at $P = 259 \ \text{hPa}$. \textbf{b}, Heat tendency at $P = 494 \ \text{hPa}$. \textbf{c}, Heat tendency at  $P = 761 \ \text{hPa}$. \textbf{d}, Moisture tendency at $P = 259 \ \text{hPa}$. \textbf{e}, Moisture tendency at $P = 494 \ \text{hPa}$. \textbf{f}, Moisture tendency at  $P = 761 \ \text{hPa}$.}
\label{fig:uncert_long_lat_all_SF}
\end{figure*}

Figures \ref{fig:uncert_long_lat_all} shows the longitude-latitude structures of the MF-RPN's uncertainty $\sigma_M$ and the MAE metric evaluated on the testing dataset. It includes the results for heat and moisture tendencies at vertical levels $259$, $494$ and $761 \ \text{hPa}$. Figures \ref{fig:uncert_long_lat_all_LF} and \ref{fig:uncert_long_lat_all_SF} provide the same results for the LF-RPN and SF-HF-RPN models respectively. Unlike the density plots and as conducted for the MAE estimates to obtain the longitude-latitude structures, the model's uncertainty $\sigma_M$ is only averaged over time here in order to be able to have a longitude-latitude visualization. Hence the standard deviation considered always corresponds to the inherent one returned by the model and is not estimated by accounting for the temporal variance.

For the heat tendency, the longitude-latitude structure of the uncertainties returned by LF-RPN and MF-RPN follows relatively well the MAE variation in longitude-latitude directions with higher values around the tropics for both metrics and also occurring in the same regions (figures \ref{fig:uncert_long_lat_all}.a - \ref{fig:uncert_long_lat_all}.c and \ref{fig:uncert_long_lat_all_LF}.a - \ref{fig:uncert_long_lat_all_LF}.c). The corresponding uncertainty variations are nearly duplicates of the MAE longitude-latitude structures. However, we can clearly notice deviations in the uncertainty's longitude-latitude structure compared to the MAE variations for the SF-HF-RPN (figures \ref{fig:uncert_long_lat_all_SF}.a - \ref{fig:uncert_long_lat_all_SF}.c). Indeed, the latter tends to overestimate the uncertainty in the tropics and temperate zone for the vertical level $259 \ \text{hPa}$ (figure \ref{fig:uncert_long_lat_all_SF}.a), while it underestimates the uncertainty in the same regions for the vertical level $761 \ \text{hPa}$ (figure \ref{fig:uncert_long_lat_all_SF}.c).

For the moisture tendency,  there is also a good agreement between the longitude-latitude structures of the uncertainty and of the MAE returned by the MF-RPN and LF-RPN models for the vertical level at $259 \ \text{hPa}$ (figures \ref{fig:uncert_long_lat_all}.d and \ref{fig:uncert_long_lat_all_LF}.d). For the moisture tendency at vertical levels $494$ and $761 \ \text{hPa}$, we still have some good agreement between the uncertainty and the MAE variations while more significant discrepancies are observed mostly within the tropics and temperate zone (figures \ref{fig:uncert_long_lat_all}.e - \ref{fig:uncert_long_lat_all}.f and \ref{fig:uncert_long_lat_all_LF}.e - \ref{fig:uncert_long_lat_all_LF}.f). The SF-HF-RPN's uncertainty also shows more important mismatches for the moisture tendency compared to the model's MAE for the vertical levels $494$ and $761 \ \text{hPa}$ compared to the $259 \ \text{hPa}$ level (figures \ref{fig:uncert_long_lat_all_SF}.d - \ref{fig:uncert_long_lat_all_SF}.f).

\begin{figure*}[ht]
\centering
\includegraphics[width=\linewidth]{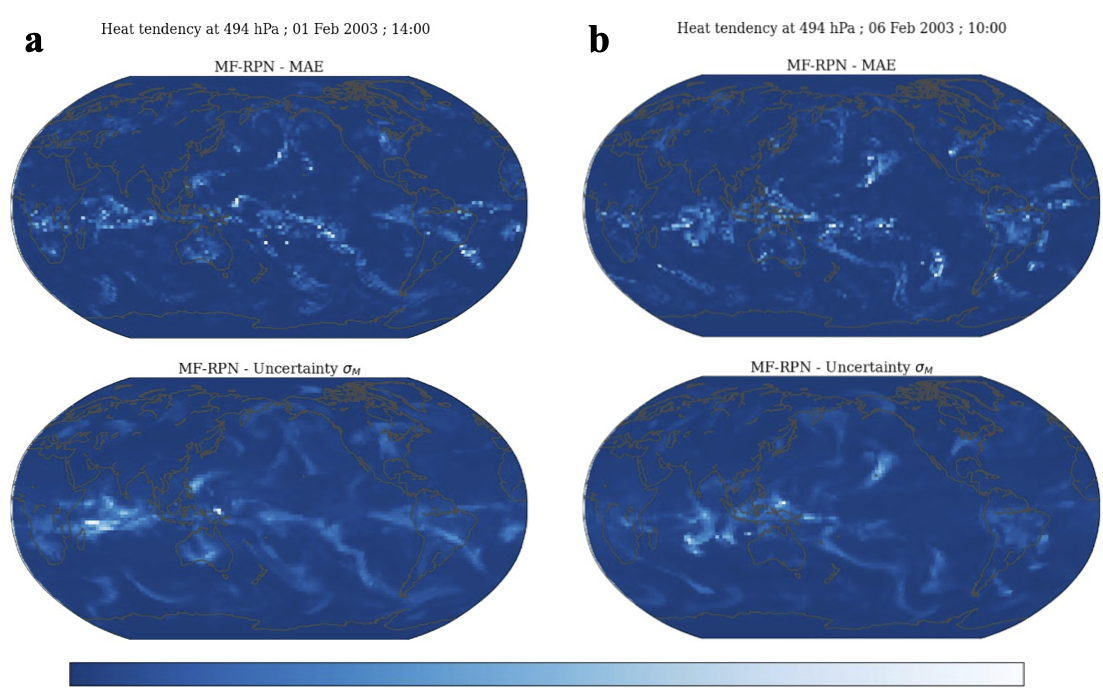}
\caption{\textbf{Longitude-latitude variation of testing MAE and quantified uncertainty $\mathbf{\sigma_M}$ at given time instances for MF-RPN for heat tendency at vertical level $P = 494 \ \text{hPa}$}. \textbf{a}, on Feb 1st 2003 at 14:00. \textbf{b}, on Feb 6th 2003 at 10:00.}
\label{fig:uncert_snapshots}
\end{figure*}

Investigating the instantaneous variations (at the hourly time-step defining the testing data) of the MAE and returned uncertainty by the MF-RPN model also shows a nearly perfect agreement (figure \ref{fig:uncert_snapshots}). For instance, on February 1st 2003 at 14:00,  all regions of high MAE values (south Atlantic ocean; south America; Pacific ocean; Australia and tropics within Africa and Indian ocean) are nearly perfectly replicated by the returned uncertainty $\mathbf{\sigma_M}$ by MF-RPN (figure \ref{fig:uncert_snapshots}.a). The same property is also observed at other time-steps, for instance on February 6th 2003 at 10:00, with the regions of high MAE values being different compared to the previous time-step, and the regions of high uncertainty $\mathbf{\sigma_M}$ shifting accordingly to well capture the regions of high MAE values (figure \ref{fig:uncert_snapshots}.b). Simulations of complete spatio-temporal evolution of MAEs and returned uncertainties for the heat and moisture tendencies by different models (MF-RPN, LF-RPN adn SF-HF-RPN) at vertical levels $259$, $494$ and $761 \ \text{hPa}$ can be found in \url{https://drive.google.com/drive/folders/1-RVxI_YpES0zNi6oa-0wIh8N1i6qL74Y?usp=drive_link}. In addition to the instantaneous variations, the simulations include evolution of daily averaged predictions and uncertainties. All simulations show the ability of the MF-RPN to return uncertainties that nearly perfectly replicate the corresponding longitude-latitude error structure at all time instances and spatial points.

Overall, the MF-RPN model returns more trustworthy uncertainty quantification compared to the LF-RPN and SF-HF-RPN models. Hence, aggregating different datasets with different fidelity levels and of different climates through a well-designed multi-fidelity surrogate model does not only improve the parameterization's deterministic and statistical error metrics, but also provides more accurate uncertainty estimates. It is worth emphasizing again that the quantified uncertainty $\sigma_M$ is estimated only based on the parameterization input without any knowledge of the true target value. Therefore, an accurate uncertainty quantification is a powerful tool, which can inform on how accurate the surrogate model is. This characteristic can be seen as similar to {\it a-posteriori} error estimates in model order reduction techniques.

\end{document}